# Extracting Patient History from Clinical Text: A Comparative Study of Clinical Large Language Models

Hieu Nghiem, MS[a,b], Tuan-Dung Le, MS[c,d], Suhao Chen, PhD[e], Thanh Thieu, PhD[d,f], Andrew Gin, MD[b], Ellie Phuong Nguyen, PhD[g], Dursun Delen, PhD[b,h,i], Johnson Thomas, PhD[a], Jivan Lamichhane, MD[j], Zhuqi Miao, PhD[k,*]

[a]Department of Computer Science, Oklahoma State University, Stillwater, OK, 74078, USA
[b]Center for Health Systems Innovation, Oklahoma State University, Stillwater, OK, 74078, USA
[c]Department of Computer Science and Engineering, University of South Florida, Tampa, FL, 33620, USA
[d]Department of Machine Learning, Moffitt Cancer Center and Research Institute, Tampa, FL, 33612, USA
[e]Department of Industrial Engineering, South Dakota School of Mines and Technology, Rapid City, SD, 57701, USA
[f]Department of Oncological Sciences, University of South Florida Morsani College of Medicine, Tampa, FL, 33612, USA
[g]South University School of Pharmacy, Savannah, GA, 31406, USA
[h]Department of Management Science and Information Systems, Oklahoma State University, Stillwater, OK, 74078, USA
[i]Department of Industrial Engineering, Faculty of Engineering and Natural Sciences, Istinye University, Sariyer/Istanbul 34396, Turkey
[j]Department of Medicine, The State University of New York Upstate Medical University, Syracuse, NY, 13210, USA
[k]School of Business, The State University of New York at New Paltz, New Paltz, NY, 12561, USA

*Corresponding Author: Zhuqi Miao, Email: miaoz@newpaltz.edu

## Abstract

**Introduction**: Extracting medical history entities (MHEs) related to a patient's chief complaint (CC), history of present illness (HPI), and past, family, and social history (PFSH) helps structure free-text clinical notes into standardized EHRs, streamlining downstream tasks like continuity of care, medical coding, and quality metrics. Fine-tuned clinical large language models (cLLMs) can assist in this process while ensuring the protection of sensitive data via on-premises deployment. This study evaluates the performance of cLLMs in recognizing CC/HPI/PFSH-related MHEs and examines how note characteristics impact model accuracy.

**Methods**: We annotated 1,449 MHEs across 61 outpatient-related clinical notes from the MTSamples repository. To recognize these entities, we fine-tuned seven state-of-the-art cLLMs. Additionally, we assessed the models' performance when enhanced by integrating, problems, tests, treatments, and other basic medical entities (BMEs). We compared the performance of these models against GPT-4o in a zero-shot setting. To further understand the textual characteristics affecting model accuracy, we conducted an error analysis focused on note length, entity length, and segmentation.

**Results**: The cLLMs showed potential in reducing the time required for extracting MHEs by over 20%. However, detecting many types of MHEs remained challenging due to their polysemous nature and the frequent involvement of non-medical vocabulary. Fine-tuned GatorTron and GatorTronS, two of the most extensively trained cLLMs, demonstrated the highest performance. Integrating pre-identified BME information improved model performance for certain entities. Regarding the impact of textual characteristics on model performance, we found that longer entities were harder to identify, note length did not correlate with a higher error rate, and well-organized segments with headings are beneficial for the extraction.

**Conclusion**: We evaluated state-of-the-art cLLMs for recognizing MHEs related to patient's CC, HPI, and PFSH. Future improvements in cLLM performance can focus on developing ensemble models that effectively integrate diverse natural language processing techniques in the biomedical domain.

## Introduction

Documenting, organizing, and analyzing clinical notes constitute a consistent challenge to healthcare professionals. The literature reveals that physicians in the U.S. spend 17% to 43% of their time on electronic health record (EHR) systems for documentary tasks [1–3], which distracts their attentions from their primary responsibility of caring for patients, and considerably lowering their productivity and career satisfaction [4]. Natural language processing (NLP), a collection of techniques that allow computers to parse and interpret human languages, provides a highly potential approach to tackle the challenge [5,6]. A subdomain of NLP—named entity recognition (NER)—focuses on extracting data elements from free text, and mapping them to specific concepts in an automatic and scalable manner [7]. In medical NLP applications, NER is often used to transform patient information embedded within unstructured narratives into structured records [8–10]. As structured EHRs are searchable and linkable (to other data sources) and can be easily analyzed, visualized and reported, this transformation facilitates a variety of downstream tasks in healthcare, such as disease prediction [11], care quality assessment [12,13], and medical coding [14,15], to improve clinical decision making [16].

One of the most prominent advances in NLP in recent years is the creation of the *Transformer* architecture [17]. Transformer-based, generative large language models (LLMs), such as ChatGPT [18], trained on diverse and extensive corpora, have demonstrated exceptional success in many NLP applications [19]. Additionally, recent research showed that in highly specialized domains like clinical note processing, clinical large language models (cLLMs), trained and fine-tuned on smaller but more specialized biomedical corpora can achieve comparable or event better performance than generative LLMs [20,21]. As far as the NER subdomain is considered, cLLMs have been extensively employed to recognize basic medical entities (BMEs), such as problems, body parts, tests, procedures, and medications, from a variety of clinical note corpora [22–25]. The applications of cLLMs to detecting patient's protected health information [26,27], disease-specific terminologies [28–30], and social/behavioral entities [31] are also frequently reported in the NLP literature.

Nevertheless, studies evaluating contemporary cLLMs in recognizing patients' detailed medical history entities (MHEs), related to their chief complaint (CC), history of present illness (HPI), and past medical, family and social history (PFSH), remain limited to date. These historical data not only assist physicians in diagnosing patients and devising treatment plans [32–35] but also foster communication among healthcare professionals to ensure consistent patient care over time [36,37]. Considering their integral roles in providing quality care, the American Medical Association requires medically appropriate history to be documented as part of the billing process for service reimbursement [38]. Hence, there is a need to investigate how cLLMs can efficiently extract and organize patient historical data.

**Objectives**: (1) Comparing the performance of state-of-the-art cLLMs fine-tuned for identifying detailed CC/HPI/PFSH-related MHEs from free-text clinical notes. (2) Investigating the textual characteristics of notes that affect the accuracy of the models through error analysis.

## Methods

### Definitions of CC, HPI and PFSH

The 1997 Documentation Guidelines for Evaluation and Management Services (1997 E/M Guidelines) provided comprehensive and pragmatic definitions for CC, HPI, and PFSH concepts [39]. We used these definitions to annotate and recognize related MHEs. According to the guidelines, CC refers to the reason for a clinical encounter, often stated in patients' own words. HPI expands on CC by providing a detailed chronological narrative of the current illness. It includes eight sub-concepts: *quality, location, severity, duration, timing, context, modifying factors,* and *associated signs and symptoms.*

Regarding PFSH, past medical history involves the patient's past experiences with illnesses, injuries, operations, medications, and allergies among others. Family history deals with a review of medical events in the patient's family, mainly about the diseases that can be hereditary or pose a risk to the patient. Social history is an age-appropriate review of the patient's past and current activities, such as marital, living, and employment statuses, alcohol or drug usages, exercises, and hobbies. Table S1 in Supplementary Material-1 provides detailed definitions and examples for each of the concepts.

## Clinical Notes

We used the Medical Transcription Sample Reports and Examples (MTSamples) as our data source. MTSamples is an open-access, de-identified clinical note repository widely utilized within the medical and medical informatics research communities [40]. The structure of MTSamples notes is very similar to that shown in Figure 1: Each note is organized as several sections with each section starting with a heading, followed by free-text narratives about the patient visit.

**HISTORY OF PRESENT ILLNESS**: A 49-year-old female with history of atopic dermatitis (past history) comes to the clinic with complaints of left otalgia (CC) and headache (CC). Symptoms started approximately three weeks (HPI duration) ago and she was having difficulty hearing (HPI sign & symptom), although that has greatly improved (HPI Quality). She is having some left-sided sinus pressure (HPI sign & symptom) and actually went to the dentist because her teeth were hurting (HPI sign & symptom); however, the teeth were okay (HPI sign & symptom). She continues to have some left-sided jaw pain (HPI sign & symptom). Denies any fever, cough, or sore throat. She had used Cutivate cream in the past for the atopic dermatitis (past history) with good results and is needing a refill of that. She has also had problems with sinusitis (past history) in the past and chronic left-sided headache (past history).

**FAMILY HISTORY**: Reviewed and unchanged.

**ALLEGIES**: To cephalexin.

**CURRENT MEDICATION**: Ibuprofen.

**SOCIAL HISTORY**: She is a nonsmoker (social history).

Figure 1. An instance of MTSamples notes used for our model training and evaluation.

We developed a benchmark corpus by annotating 1,449 MHEs across 61 outpatient-related clinical notes from MTSamples. These notes comprised various types, including 27 consultation notes, 12 subjective, objective, assessment, and plan (SOAP) reports, 6 emergency room reports, 3 follow-up notes, 3 history and physical notes, and 10 miscellaneous notes, totaling 3,578 sentences and 44,385 tokens. Table 1-Part (a) provides a detailed count of MHEs for each concept. The annotations were first completed by two undergraduate biology major students in a double-blind manner, and then curated by two co-authors (Z.M. and S.C.), followed by verification and adjustment by another co-author (A.G.), an experienced doctor.

Table 1. The counts of MHEs and BMEs included in our corpus.

| Part (a) | | Part (b) | |
|---|---|---|---|
| MHE | Count | BME (pre-identified by CLAMP) | Count |
| CC | 133 | Problem | 3,111 |
| HPI Location | 48 | Test | 1,007 |
| HPI Quality | 53 | Treatment | 479 |
| HPI Severity | 34 | Drug | 481 |
| HPI Duration | 66 | Body Location | 973 |
| HPI Timing | 37 | Severity | 109 |
| HPI Context | 37 | Temporal | 581 |
| HPI Modifying Factor | 81 | | |
| HPI Associated Sign and Symptom | 268 | | |
| Past Medical History | 518 | | |
| Family History | 45 | | |
| Social History | 129 | | |

### CLLMs and Fine-Tuning

We used OpenAI's latest flagship model, GPT-4o [41], in a zero-shot setting [42], as the base-line. In the modeling, GPT-4o was instructed to extract the MHEs and generate the output in HTML format via OpenAI's API. The MHEs are marked up with a *"<span>"* tag, and each *"<span>"* tag has a class attribute indicating the type of the entity (detailed prompt is available in Supplementary Material-2).

Then, we fine-tuned seven cLLMs, including Bio+Discharge Summary BERT, Bio+Clinical BERT [43], PubMedBERT [44], PubMedBERT Large [45], BioMegatron [46], GatorTron [47], and GatorTronS [48]. Two finetuning approaches were employed. The first approach was a basic fine-tuning scheme utilizing a fully-connected linear output layer [49]. As illustrated in Figure 2. First, we converted the MHE annotations into BIO tags: "B-<MHE>" for tokens (words or punctuations) at the beginning, "I-<MHE>" for tokens inside, and "O" for tokens outside a specific <MHE> (<MHE> represents a placeholder that was replaced with the specific MHE type indicated by the annotation). The output layer then assigns the BIO tag to each token, enabling the recognition of entities.

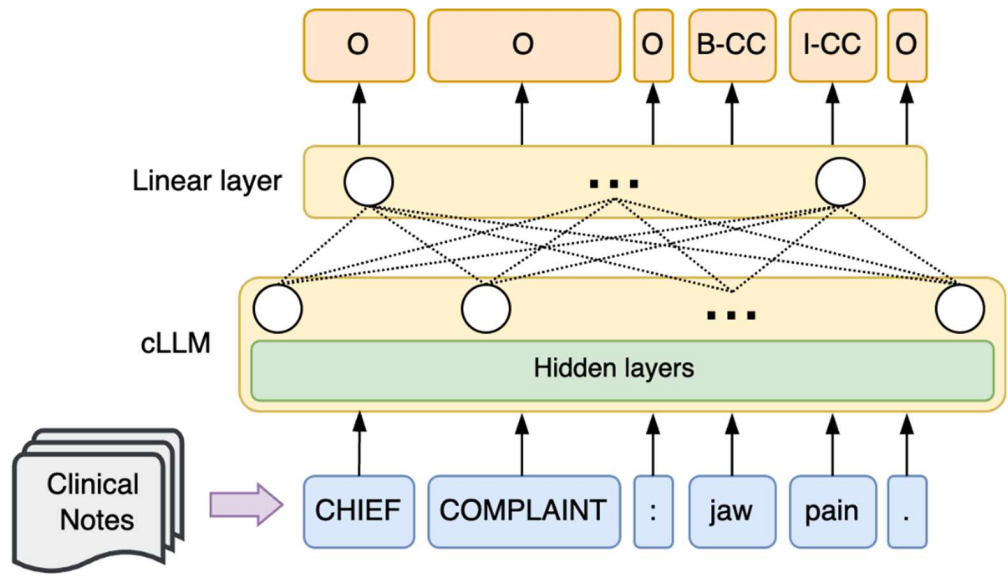

Figure 2. The basic fine-tuning architecture using a fully connected output layer.

The second approach expanded upon the first by incorporating pre-identified BME information as clues, aiming to enhance the downstream accuracy of recognizing MHEs. The specific types of BMEs we used are listed in Table 1-Part (b) and were pre-identified using the Clinical Language Annotation, Modeling, and Processing Toolkit (CLAMP) [50]. As illustrated in Figure 3, we first obtained the BIO tags of the BMEs, which were then one-hot encoded and concatenated with the cLLM embeddings. The concatenated embeddings served as input to the output layer to classify the MHE BIO tags.

For both fine-tuning approaches, we employed five-fold cross-validation for the modeling and the evaluation over the benchmark corpus [51]. Additional details regarding the cLLMs and the two fine-tuning approaches are provided in Supplementary Material-2.

### Model Evaluation

Our evaluation of the fine-tuned cLLMs focused on two aspects: accuracy assessment and error association analysis, as discussed below.

**Accuracy Assessment**: We considered five types of matching scenarios between cLLM detections and ground truth annotations to assess the model accuracy [52]:

- *Exact Match*: It is subject to a strict criterion that considers a prediction correct if and only if both the predicted concept and entity span match exactly to the annotation.

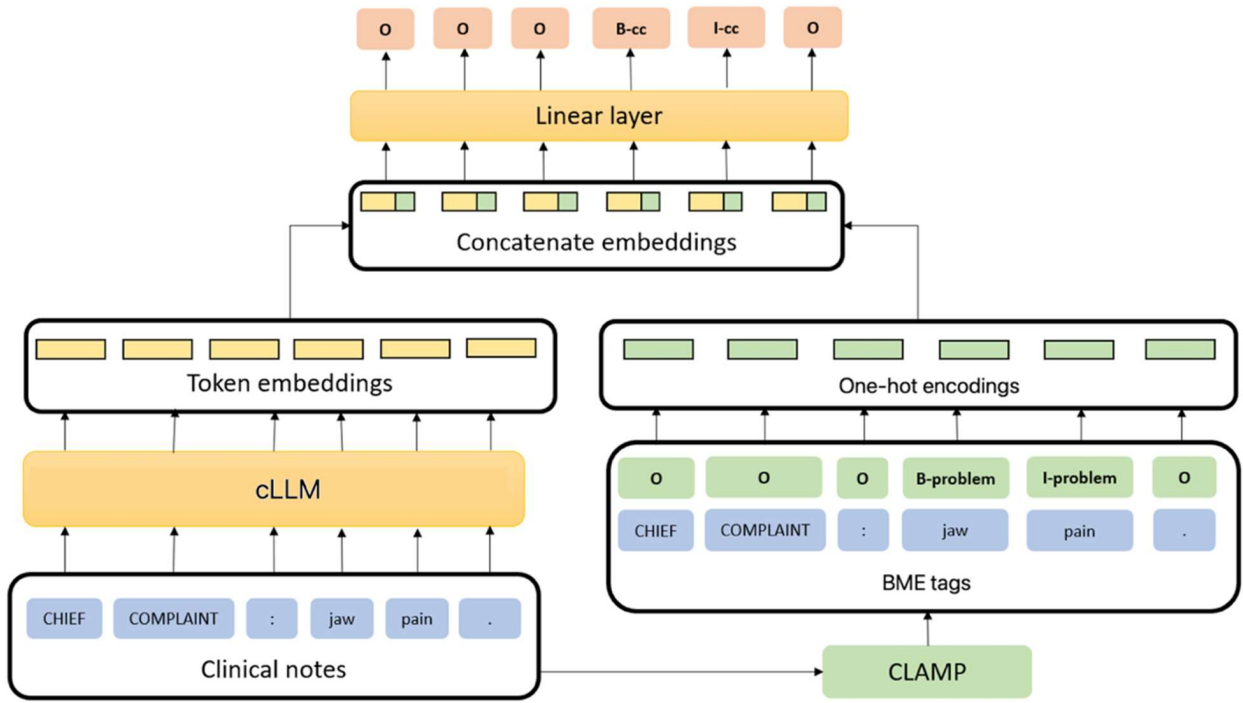

Figure 3. The fine-tuning architecture incorporating BMEs pre-identified by CLAMP.

- *Relaxed Match*: It accepts "partially correct" prediction, where the concept is predicted correctly, while the entity span, even if does not precisely match but overlaps with the annotation.
- *Mismatch*: The predicted concept does not match the annotation even though there is overlap in the entity span.
- *Under Detection*: The model fails to detect an entity overlapping with the annotation.
- *Over Detection*: Misidentification of text not annotated as an MHE.

Mismatches and under detections (MMUD) are entities that were recognized improperly and are therefore of interest in our error analysis. MMUD and over detections are considered as errors. The error rate, relative to the total number of annotations, is a key measure as it reflects the frequency of corrections by a human-end user utilizing this system in a medical workflow.

**Error Association Analysis:** The textual characteristics of notes investigated and the statistical tests used in our error association analysis are outlined below. All tests were conducted at a 0.05 significance level using R.

- *Entity Length*. We hypothesized that entities with more tokens are less likely to match exactly. We evaluated the hypothesis by comparing the entity lengths of exact matches against those of each other matching category using two-sample *t*-tests.
- *Note Length*. It is intuitive that longer notes (measured by the number of tokens) contain more entities to identify, potentially resulting in both more matches and errors. We verified it using correlation tests. Moreover, we hypothesized that lengthy notes would not increase the error rate. This hypothesis was also tested using correlation test.
- *Segmentation*. We hypothesized that entities located in dedicated sections with explicit headers are easier to identify. To test this hypothesis, we compared the occurrence of MMUD between entities found within sections with correct headers and those found elsewhere, using $\chi^2$-tests.

## Results

### Model Accuracy

In Table 2, the counts of exact matches, relaxed matches, and errors are presented for each model. The results indicate that GatorTron+CLAMP and GatorTronS+CLAMP stood out with the lowest error rates, at 61.4% and 61.8%, respectively. Relaxed matches may also require corrections though likely involving less efforts. Considering the rate of the entities needing correction (relaxed matches + errors), GatorTronS+CLAMP outperformed the other models, achieving the lowest rate of 75.9% (14.1% + 61.8%). It suggests that using the model may save over 20% of the time on tasks extracting MHEs. Interestingly, the fine-tuned cLLMs significantly outperformed the baseline, zero-shot GPT-4o model, similar to recent reports of GPT's performance on BMEs in the literature [42]. It suggests that zero-shot learning is not yet adequate for handling the complexity of MHEs related to CC/HPI/PFSH. Notably, the model demonstrated a high rate of over-detection.

Table 2: The count and rate of matching cLLMs prediction against ground truth annotation.

|  | Exact Match | Relaxed Match | Errors | | | |
|---|---|---|---|---|---|---|
|  |  |  | Mismatch | Under Detection | Over Detection | Total |
| OpenAI GPT-4o with zero-shot learning | 363 (25.1%) | 222 (15.3%) | 191 (13.2%) | 689 (47.6%) | 1498 (103.4%) | 2378 (164.1%) |
| Fine-Tuned Models; Without CLAMP | | | | | | |
| Bio+Discharge Summary BERT | 672 (46.4%) | 212 (14.6%) | 156 (10.8%) | 433 (29.9%) | 462 (31.9%) | 1051 (72.5%) |
| Bio+Clinical BERT | 661 (45.6%) | 208 (14.4%) | 176 (12.1%) | 434 (30.0%) | 482 (33.3%) | 1092 (75.4%) |
| BioMegatron | 670 (46.2%) | 209 (14.4%) | 165 (11.4%) | 441 (30.4%) | 477 (32.9%) | 1083 (74.7%) |
| PubMedBERT | 679 (46.9%) | 180 (12.4%) | 183 (12.6%) | 427 (29.5%) | 500 (34.5%) | 1110 (76.6%) |
| PubMedBERT Large | 671 (46.3%) | 200 (13.8%) | 173 (11.9%) | 436 (30.1%) | 459 (31.7%) | 1068 (73.7%) |
| GatorTron | 746 (51.5%) | 229 (15.8%) | 142 (9.8%) | 364 (25.1%) | 395 (27.3%) | 901 (62.2%) |
| GatorTronS | 719 (49.6%) | 209 (14.4%) | 120 (8.3%) | 433 (29.9%) | 401 (27.7%) | 954 (65.8%) |
| Fine-Tuned Models; With CLAMP | | | | | | |
| Bio+Discharge Summary BERT | 665 (45.9%) | 206 (14.2%) | 160 (11.0%) | 448 (30.9%) | 444 (30.6%) | 1052 (72.6%) |
| Bio+Clinical BERT | 661 (45.6%) | 217 (15.0%) | 185 (12.8%) | 428 (29.5%) | 525 (36.2%) | 1138 (78.5%) |
| BioMegatron | 642 (44.3%) | 202 (13.9%) | 142 (9.8%) | 492 (34.0%) | 445 (30.7%) | 1079 (74.5%) |
| PubMedBERT | 639 (44.1%) | 194 (13.4%) | 182 (12.6%) | 469 (32.4%) | 474 (32.7%) | 1125 (77.6%) |
| PubMedBERT Large | 665 (45.9%) | 183 (12.6%) | 150 (10.4%) | 480 (33.1%) | 430 (29.7%) | 1060 (73.2%) |
| GatorTron | 720 (49.7%) | 232 (16.0%) | 129 (8.9%) | 409 (28.2%) | 351 (24.2%) | ***889 (61.4%)*** |
| GatorTronS | 698 (48.2%) | 204 (14.1%) | 114 (7.9%) | 459 (31.7%) | 323 (22.3%) | ***896 (61.8%)*** |

Note: The rate is calculated as the ratio of the number of predictions to the total number of annotations. Since a single annotated entity can sometimes overlap multiple predicted entities, the summation of ratios for exact match, relaxed match, and error may exceed one.

Given that GatorTron/GatorTronS-based cLLMs exhibit better performance, we focus on presenting their results in the remainder of this article. Table 3 shows the error rates of GatorTron/GatorTronS-based models for each CC/HPI/PFSH concept. Other models' error rates are available in Supplementary Material-3, Table S2. They reveal that:

- Most MHEs are difficult to detect, with error rates frequently exceeding 50%. The challenge may arise from several factors:
  - CC/HPI/PFSH concepts can be polysemous, meaning a single concept can encompass a broad range of sub-concepts, which may confuse cLLMs. For instance, a modifying factor can be a medication, an environmental factor, or an action.

Table 3. Error rates of GatorTron/GatorTronS-based models for each CC/HPI/PFSH concept.

| | GatorTron | | | | GatorTron + CLAMP | | | |
|---|---|---|---|---|---|---|---|---|
| | Mismatch | Under Detection | Over Detection | Total Errors | Mismatch | Under Detection | Over Detection | Total Errors |
| CC | 38 (28.6%) | 16 (12.0%) | 14 (10.5%) | 68 (51.1%) | 31 (23.3%) | 19 (14.3%) | 20 (15.0%) | 70 (52.6%) |
| Location | 7 (14.6%) | 16 (33.3%) | 10 (20.8%) | 33 (68.8%) | 7 (14.6%) | 16 (33.3%) | 10 (20.8%) | 33 (68.8%) |
| Quality | 5 (9.4%) | 19 (35.8%) | 14 (26.4%) | 38 (71.7%) | 4 (7.5%) | 20 (37.7%) | 13 (24.5%) | **37 (69.8%)*** |
| Severity | 9 (26.5%) | 16 (47.1%) | 9 (26.5%) | 34 (100.0%) | 4 (11.8%) | 20 (58.8%) | 10 (29.4%) | 34 (100.0%) |
| Duration | 1 (1.5%) | 12 (18.2%) | 24 (36.4%) | 37 (56.1%) | 1 (1.5%) | 12 (18.2%) | 21 (31.8%) | **34 (51.5%)** |
| Timing | 9 (24.3%) | 8 (21.6%) | 12 (32.4%) | 29 (78.4%)* | 9 (24.3%) | 12 (32.4%) | 11 (29.7%) | 32 (86.5%) |
| Context | 13 (35.1%) | 9 (24.3%) | 10 (27.0%) | 32 (86.5%)* | 13 (35.1%) | 12 (32.4%) | 9 (24.3%) | 34 (91.9%) |
| M.F. | 10 (12.3%) | 27 (33.3%) | 30 (37.0%) | 67 (82.7%) | 8 (9.9%) | 29 (35.8%) | 26 (32.1%) | **63 (77.8%)*** |
| Symptom | 26 (9.7%) | 109 (40.7%) | 132 (49.3%) | 267 (99.6%) | 35 (13.1%) | 124 (46.3%) | 94 (35.1%) | **253 (94.4%)*** |
| Past H. | 18 (3.5%) | 113 (21.8%) | 122 (23.6%) | 253 (48.8%) | 11 (2.1%) | 123 (23.7%) | 118 (22.8%) | **252 (48.6%)** |
| Fam. H. | 0 (0.0%) | 2 (4.4%) | 6 (13.3%) | 8 (17.8%)* | 0 (0.0%) | 2 (4.4%) | 6 (13.3%) | 8 (17.8%)* |
| Social H. | 6 (4.7%) | 17 (13.2%) | 12 (9.3%) | 35 (27.1%) | 6 (4.7%) | 20 (15.5%) | 13 (10.1%) | 39 (30.2%) |
| | GatorTronS | | | | GatorTronS + CLAMP | | | |
| | Mismatch | Under Detection | Over Detection | Total Errors | Mismatch | Under Detection | Over Detection | Total Errors |
| CC | 25 (18.8%) | 30 (22.6%) | 24 (18.0%) | 79 (59.4%) | 24 (18.0%) | 22 (16.5%) | 19 (14.3%) | **65 (48.9%)*** |
| Location | 8 (16.7%) | 17 (35.4%) | 10 (20.8%) | 35 (72.9%) | 8 (16.7%) | 19 (39.6%) | 5 (10.4%) | **32 (66.7%)*** |
| Quality | 4 (7.5%) | 21 (39.6%) | 14 (26.4%) | 39 (73.6%) | 7 (13.2%) | 25 (47.2%) | 9 (17.0%) | 41 (77.4%) |
| Severity | 8 (23.5%) | 16 (47.1%) | 6 (17.6%) | 30 (88.2%)* | 6 (17.6%) | 25 (73.5%) | 2 (5.9%) | 33 (97.1%) |
| Duration | 0 (0.0%) | 18 (27.3%) | 21 (31.8%) | 39 (59.1%) | 0 (0.0%) | 19 (28.8%) | 14 (21.2%) | **33 (50.0%)*** |
| Timing | 9 (24.3%) | 14 (37.8%) | 10 (27.0%) | 33 (89.2%) | 8 (21.6%) | 15 (40.5%) | 9 (24.3%) | **32 (86.5%)** |
| Context | 13 (35.1%) | 13 (35.1%) | 12 (32.4%) | 38 (102.7%) | 15 (40.5%) | 15 (40.5%) | 8 (21.6%) | 38 (102.7%) |
| M.F. | 7 (8.6%) | 39 (48.1%) | 32 (39.5%) | 78 (96.3%) | 9 (11.1%) | 35 (43.2%) | 28 (34.6%) | **72 (88.9%)** |
| Symptom | 27 (10.1%) | 120 (44.8%) | 125 (46.6%) | 272 (101.5%) | 23 (8.6%) | 132 (49.3%) | 110 (41.0%) | **265 (98.9%)** |
| Past H. | 14 (2.7%) | 124 (23.9%) | 126 (24.3%) | 264 (51.0%) | 12 (2.3%) | 117 (22.6%) | 101 (19.5%) | **230 (44.4%)*** |
| Fam. H. | 1 (2.2%) | 4 (8.9%) | 5 (11.1%) | 10 (22.2%) | 0 (0.0%) | 4 (8.9%) | 5 (11.1%) | 9 (20.0%) |
| Social H. | 4 (3.1%) | 17 (13.2%) | 16 (12.4%) | 37 (28.7%) | 2 (1.6%) | 31 (24.0%) | 13 (10.1%) | 46 (35.7%) |

Note: The **Bold** font highlights the cases where the integration of CLAMP improves accuracy; * indicates the lowest error rate among all models assessed for each concept. Abbreviations: M.F. (modifying factors), Past H. (past medical history), Fam. H. (family history), Social H. (social history).

  - Many entities are primarily composed of non-medical vocabulary. Such entities are frequently associated with concepts like HPI-Context and HPI-Timing, where error rates are high. In contrast, family history primarily involving conditions among a patient's family members were easier to be detected correctly, with at lowest 17.8% error rate.

Examples regarding the challenges are provided in Supplementary Material-1.

- The inclusion of CLAMP inputs frequently improved the accuracy in detecting CC, location, duration, and modifying factors for at least 5 out of the 7 fine-tuned cLLMs (see Supplementary Material-3, Table S2). The enhancement might be attributed to CLAMP providing problem information, which is central to the CC concept. Additionally, the body location, temporal, and medication tags from CLAMP may assist in identifying HPI location, duration, and modifying factor entities.

## Error Analysis

Table 4 compares the average entity lengths across all the five categories of matching results for GatorTron/GatorTronS-based cLLMs. The results indicate that exact matches tend to be short entities. Interestingly, irrelevant text of similar length to exact matches may confuse the models, leading to over detections. Longer entities, on the other hand, are more prone to relaxed matches, mismatches, and under detections.

Table 4. The average and standard deviation (SD) of entity lengths for each matching category and their associated two-sample *t*-test results for GatorTron/GatorTronS-based cLLMs.

|  | GatorTron | | | GatorTron+CLAMP | | |
| --- | --- | --- | --- | --- | --- | --- |
|  | Avg | SD | *p*-value | Avg | SD | *p*-value |
| Exact Match | 1.952 | 1.512 | reference | 1.821 | 1.354 | reference |
| Relaxed Match | 4.670 | 3.766 | <0.001 | 4.712 | 3.553 | <0.001 |
| Mismatch | 3.028 | 2.873 | <0.001 | 2.915 | 2.434 | <0.001 |
| Under Detection | 3.088 | 2.983 | <0.001 | 3.284 | 3.200 | <0.001 |
| Over Detection | 2.126 | 1.933 | 0.109 | 2.003 | 1.539 | 0.049 |
|  | GatorTronS | | | GatorTronS+CLAMP | | |
|  | Avg | SD | *p*-value | Avg | SD | *p*-value |
| Exact Match | 1.846 | 1.506 | reference | 1.809 | 1.394 | reference |
| Relaxed Match | 4.887 | 4.246 | <0.001 | 4.596 | 3.651 | <0.001 |
| Mismatch | 3.175 | 2.649 | <0.001 | 2.772 | 2.566 | <0.001 |
| Under Detection | 3.134 | 2.640 | <0.001 | 3.340 | 3.060 | <0.001 |
| Over Detection | 1.813 | 1.678 | 0.740 | 1.827 | 1.433 | 0.848 |

Figure 4(A) shows that as note length increases, there is a statistically significant rise in the count of errors made by GatorTron+CLAMP (correlation coefficient=0.615, *p*-value<0.001). This is intuitive because more tokens necessitate larger classification spaces, which in turn leads to more potential errors. However, increasing note length does not necessarily lead to a proportional rise in the error rate, as shown in Figure 4(B). The correlation coefficient is 0.034 with a *p*-value=0.796. It suggests the model's potential robustness in handling long notes without a substantial increase in error rates, which can be a favorable feature of the model in real-world applications. The results for other GatorTron/GatorTronS-based models exhibited a similar pattern as shown in Supplementary Material-3, Table S3.

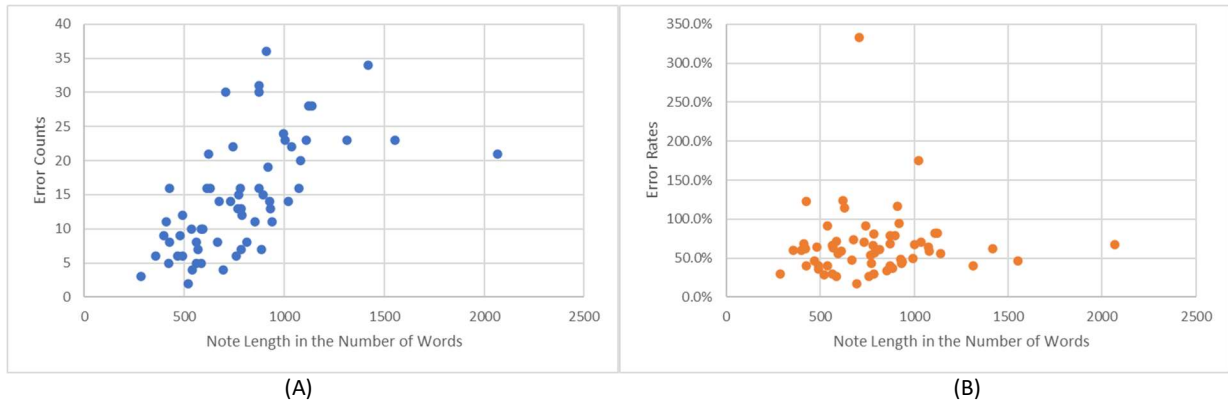

Figure 4. The correlations between note lengths and (A) error counts and (B) error rates for GatorTron+CLAMP.

Figure 5 shows that both GatorTron+CLAMP and GatorTronS+CLAMP more accurately identified the CC and PFSH entities when they are within the dedicated headed sections. The better performance is stastically significant (*p*-values<0.05). This suggests that well-organized segments and indicative headings can facilitate the recognition of CC and PFSH entities. It could be because CC and PFSH are often related to problems and medications, which, without correct headings, can be easily confused with HPI entities. However, there was no significant difference in the accuracy of recognizing HPI entities, regardless of whether the entities were in the dedicated headed sections. This might be due to the specificity of many HPI entities, like location and duration, makes them recognizable regardless of their placement or the presence of headings. The results for GatorTron and GatorTronS exhibit a similar pattern, as shown in Supplementary Material-3, Table S4.

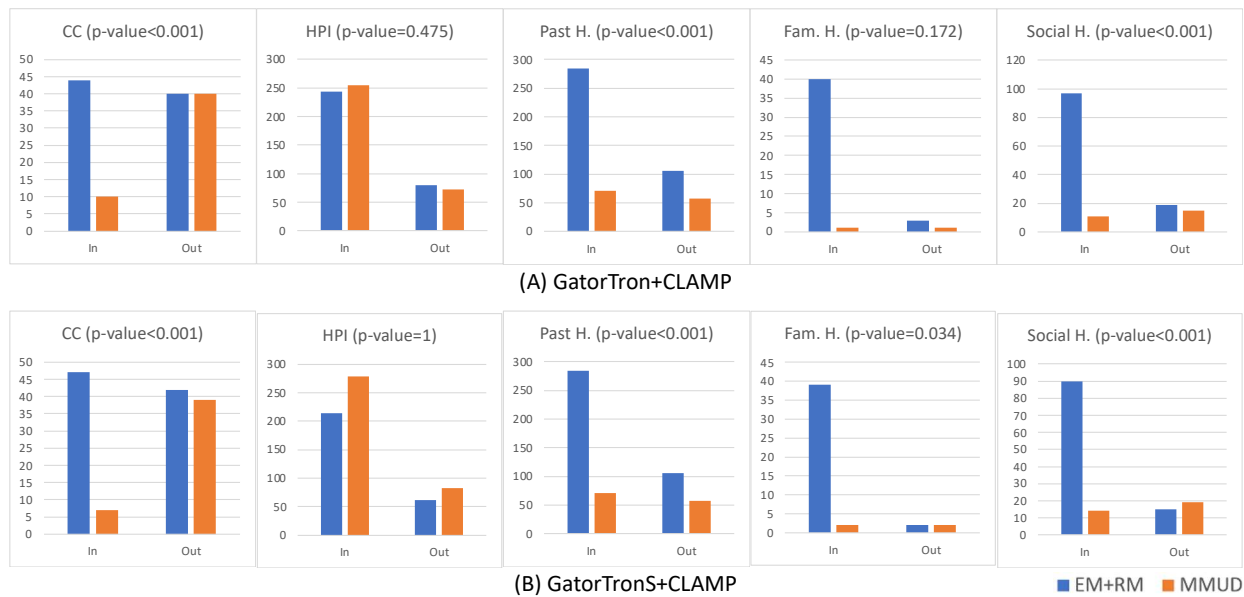

Figure 5. Comparison of occurrences of MMUD between entities inside and outside of dedicated sections with headers, and associated $\chi^2$-test *p*-values. Abbreviations: EM (exact match), RM (relaxed match).

## Discussion

Patient history plays a pivotal role in guiding clinical evaluations, decision-making, and personalized patient management [53]. Precise extraction and organized presentation of CC/HPI/PFSH-related MHEs enables quick access and review and facilitate communication among the care team members [54,55]. Furthermore, structured MHEs ease the use of statistical, machine/deep learning, and visualization tools for analysis, providing significant advantages for secondary research and quality metrics [56]. LLMs have emerged as useful tools for automating the recognition of MHEs to enhance the efficiency and scalability of the extraction process. Generative LLMs, such as ChatGPT, can be adapted to MHE recognition through few-shot learning [57]. However, leading generative LLMs are often hosted online, which poses significant risks when used for sensitive healthcare data due to potential breaches and lack of control over data privacy [58,59]. Utilizing open-source LLMs, such as Meta Llama [60], might provide some relief by allowing for on-premises deployment, making it an interesting direction for future work. Tailoring cLLMs to healthcare organizations' specific data and tasks through fine-tuning presents another promising solution. The specialized model allows on-premises deployment within the secure confines of the healthcare institution's own IT infrastructure, ensuring that patient data remains protected and complies with regulations such as HIPAA [61], while performing NLP tasks. A caveat of adopting cLLMs compared to few-shot learning with generative LLMs is that fine-tuning necessitates a training dataset, leading to significantly more annotations that are costly and time-consuming.

Our results indicate that lengthier text inputs tend to increase the number of over detections. To mitigate this, we recommend processing only the minimal text segments necessary for MHE extraction. EHR systems often structure notes using templates with predefined headings [62,63], which help identify the most relevant sections. Moreover, when notes are well-organized into designated sections, the detection of CC and PFSH becomes more accurate. Despite the strategies, extracting MHEs, especially those related to HPI, presents significant challenges. This complexity likely arises from the polysemous nature of many HPI concepts, such as modifying factors, past medical history, and associated signs and symptoms. The inclusion of a significant proportion of non-medical vocabulary within HPI context, quality, and timing, as well as other types of entities (examples are available in Supplementary Materials-1), also pose challenges, as identifying them heavily relies on understanding their semantic relationships in the clinical note context.

Therefore, there is considerable room and need to improve cLLMs in recognizing MHEs. From a technical perspective, there are three potential areas for improvement: First, we suggest developing an ensemble of different NLP techniques to handle complex MHEs. The improvement achieved by incorporating BME information identified by CLAMP in this study demonstrated the potential of this approach. A multitude of NER tools are currently available for identifying BMEs [40,64]. Exploring more efficient strategies to integrate these tools into cLLMs can be instrumental in enhancing the performance of the models.

Additionally, generative LLMs that excel at discovering semantic relationships within text [65,66] are potentially well-suited for handling entities like HPI context, quality, and timing, among others. In such cases, semantic relations play a more crucial role in recognition than individual medical terms. Second, over-detections can introduce considerable redundancy in the patient history extraction process. Applying cost-sensitive learning techniques [67] that assign penalties to over-detections may help mitigate this issue and enhance the entity recognition accuracy. Third, our evaluation indicates that long entities are particularly challenging to capture fully, often resulting in relaxed matches with fragmented or incomplete information extraction. Hence, NER techniques tailored for long textual span, though currently underdeveloped, are desired to enhance the accuracy of cLLMs in recognizing lengthy MHEs.

*Limitation*: The annotation capacity of our team focused this study on a selected sample of outpatient-related clinical notes. As a result, the performance of cLLMs on other types of notes is not yet fully explored. Future work should expand the evaluation to a broader corpus that includes a wider variety of clinical note types to better assess cLLMs' effectiveness in MHE recognition.

## Conclusion

Patient history is crucial for clinical evaluation, yet it is typically documented in clinical notes that are time-consuming to review. This study shows that cLLMs offer a highly potential solution to extract key MHEs, such as CC, HPI, and PFSH, from free text for streamlining the access to the essential information. By employing fine-tuned state-of-the-art cLLMs, such as GatorTron/GatorTronS, time for extracting MHEs can potentially be reduced by 20% compared to human processing. However, challenges remain, necessitating the creation of ensemble NER techniques to handle polysemous MHEs. Additionally, further advancements in NLP techniques, focused on mitigating over-detection and enhancing the accuracy of recognizing long entities, are needed to improve the performance of cLLMs.

# Funding

This study is funded by grant #HR21-173 through the Oklahoma Center for the Advancement of Science and Technology.

# Data Availability

The annotated clinical notes used in this study are shared publicly on GitHub at https://github.com/isallexist/Patient-History-NER.

# Code Availability

The code supporting the findings of this study is shared publicly on GitHub at https://github.com/isallexist/Patient-History-NER.

# References

[1] C. Sinsky, L. Colligan, L. Li, M. Prgomet, S. Reynolds, L. Goeders, J. Westbrook, M. Tutty, G. Blike, Allocation of Physician Time in Ambulatory Practice: A Time and Motion Study in 4 Specialties, Ann Intern Med 165 (2016) 753–760. https://doi.org/10.7326/M16-0961.

[2] B.G. Arndt, J.W. Beasley, M.D. Watkinson, J.L. Temte, W.-J. Tuan, C.A. Sinsky, V.J. Gilchrist, Tethered to the EHR: Primary Care Physician Workload Assessment Using EHR Event Log Data and Time-Motion Observations, Ann Fam Med 15 (2017) 419–426. https://doi.org/10.1370/afm.2121.

[3] M. Tai-Seale, C.W. Olson, J. Li, A.S. Chan, C. Morikawa, M. Durbin, W. Wang, H.S. Luft, Electronic Health Record Logs Indicate That Physicians Split Time Evenly Between Seeing Patients And Desktop Medicine, Health Affairs 36 (2017) 655–662. https://doi.org/10.1377/hlthaff.2016.0811.

[4] S. Woolhandler, D.U. Himmelstein, Administrative Work Consumes One-Sixth of U.S. Physicians' Working Hours and Lowers their Career Satisfaction, Int J Health Serv 44 (2014) 635–642. https://doi.org/10.2190/HS.44.4.a.

[5] P.M. Nadkarni, L. Ohno-Machado, W.W. Chapman, Natural language processing: an introduction, Journal of the American Medical Informatics Association 18 (2011) 544–551. https://doi.org/10.1136/amiajnl-2011-000464.

[6] S. Locke, A. Bashall, S. Al-Adely, J. Moore, A. Wilson, G.B. Kitchen, Natural language processing in medicine: A review, Trends in Anaesthesia and Critical Care 38 (2021) 4–9. https://doi.org/10.1016/j.tacc.2021.02.007.

[7] D. Nadeau, S. Sekine, A survey of named entity recognition and classification, Lingvisticæ Investigationes 30 (2007) 3–26. https://doi.org/10.1075/li.30.1.03nad.

[8] J.M. Giorgi, G.D. Bader, Towards reliable named entity recognition in the biomedical domain, Bioinformatics 36 (2020) 280–286. https://doi.org/10.1093/bioinformatics/btz504.

[9] P. Bose, S. Srinivasan, W.C. Sleeman, J. Palta, R. Kapoor, P. Ghosh, A Survey on Recent Named Entity Recognition and Relationship Extraction Techniques on Clinical Texts, Applied Sciences 11 (2021) 8319. https://doi.org/10.3390/app11188319.

[10] D. Fraile Navarro, K. Ijaz, D. Rezazadegan, H. Rahimi-Ardabili, M. Dras, E. Coiera, S. Berkovsky, Clinical named entity recognition and relation extraction using natural language processing of medical free text: A systematic review, International Journal of Medical Informatics 177 (2023) 105122. https://doi.org/10.1016/j.ijmedinf.2023.105122.

[11] E. Ford, J.A. Carroll, H.E. Smith, D. Scott, J.A. Cassell, Extracting information from the text of electronic medical records to improve case detection: a systematic review, Journal of the American Medical Informatics Association 23 (2016) 1007–1015. https://doi.org/10.1093/jamia/ocv180.

[12] K.C. Lee, B.V. Udelsman, J. Streid, D.C. Chang, A. Salim, D.H. Livingston, C. Lindvall, Z. Cooper, Natural Language Processing Accurately Measures Adherence to Best Practice Guidelines for Palliative Care in Trauma, Journal of Pain and Symptom Management 59 (2020) 225-232.e2. https://doi.org/10.1016/j.jpainsymman.2019.09.017.

[13] A. Turchin, L.F. Florez Builes, Using Natural Language Processing to Measure and Improve Quality of Diabetes Care: A Systematic Review, J Diabetes Sci Technol 15 (2021) 553–560. https://doi.org/10.1177/19322968211000831.

[14] K.S. Kalyan, S. Sangeetha, SECNLP: A survey of embeddings in clinical natural language processing, Journal of Biomedical Informatics 101 (2020) 103323. https://doi.org/10.1016/j.jbi.2019.103323.

[15] H. Dong, M. Falis, W. Whiteley, B. Alex, J. Matterson, S. Ji, J. Chen, H. Wu, Automated clinical coding: what, why, and where we are?, Npj Digit. Med. 5 (2022) 1–8. https://doi.org/10.1038/s41746-022-00705-7.

[16] D. Demner-Fushman, W.W. Chapman, C.J. McDonald, What can natural language processing do for clinical decision support?, Journal of Biomedical Informatics 42 (2009) 760–772. https://doi.org/10.1016/j.jbi.2009.08.007.

[17] A. Vaswani, N. Shazeer, N. Parmar, J. Uszkoreit, L. Jones, A.N. Gomez, Ł. Kaiser, I. Polosukhin, Attention is All you Need, in: Advances in Neural Information Processing Systems, Curran Associates, Inc., 2017. https://proceedings.neurips.cc/paper/2017/hash/3f5ee243547dee91fbd053c1c4a845aa-Abstract.html (accessed March 26, 2023).

[18] T. Wu, S. He, J. Liu, S. Sun, K. Liu, Q.-L. Han, Y. Tang, A Brief Overview of ChatGPT: The History, Status Quo and Potential Future Development, IEEE/CAA J. Autom. Sinica 10 (2023) 1122–1136. https://doi.org/10.1109/JAS.2023.123618.

[19] W.X. Zhao, K. Zhou, J. Li, T. Tang, X. Wang, Y. Hou, Y. Min, B. Zhang, J. Zhang, Z. Dong, Y. Du, C. Yang, Y. Chen, Z. Chen, J. Jiang, R. Ren, Y. Li, X. Tang, Z. Liu, P. Liu, J.-Y. Nie, J.-R. Wen, A Survey of Large Language Models, arXiv.Org (2023). https://arxiv.org/abs/2303.18223v13 (accessed June 4, 2024).

[20] E. Lehman, E. Hernandez, D. Mahajan, J. Wulff, M.J. Smith, Z. Ziegler, D. Nadler, P. Szolovits, A. Johnson, E. Alsentzer, Do We Still Need Clinical Language Models?, in: Proceedings of the Conference on Health, Inference, and Learning, PMLR, 2023: pp. 578–597. https://proceedings.mlr.press/v209/eric23a.html (accessed June 4, 2024).

[21] A.J. Thirunavukarasu, D.S.J. Ting, K. Elangovan, L. Gutierrez, T.F. Tan, D.S.W. Ting, Large language models in medicine, Nat Med 29 (2023) 1930–1940. https://doi.org/10.1038/s41591-023-02448-8.

[22] J.M. Steinkamp, W. Bala, A. Sharma, J.J. Kantrowitz, Task definition, annotated dataset, and supervised natural language processing models for symptom extraction from unstructured clinical notes, Journal of Biomedical Informatics 102 (2020) 103354. https://doi.org/10.1016/j.jbi.2019.103354.

[23] X. Yang, J. Bian, W.R. Hogan, Y. Wu, Clinical concept extraction using transformers, J Am Med Inform Assoc 27 (2020) 1935–1942. https://doi.org/10.1093/jamia/ocaa189.

[24] J. Li, Q. Wei, O. Ghiasvand, M. Chen, V. Lobanov, C. Weng, H. Xu, A comparative study of pre-trained language models for named entity recognition in clinical trial eligibility criteria from multiple corpora, BMC Medical Informatics and Decision Making 22 (2022) 235. https://doi.org/10.1186/s12911-022-01967-7.

[25] S. Narayanan, M. S.s., M.V. Ramesh, P.V. Rangan, S.P. Rajan, Deep contextual multi-task feature fusion for enhanced concept, negation and speculation detection from clinical notes, Informatics in Medicine Unlocked 34 (2022) 101109. https://doi.org/10.1016/j.imu.2022.101109.

[26] S.H. Oh, M. Kang, Y. Lee, Protected Health Information Recognition by Fine-Tuning a Pre-training Transformer Model, Healthc Inform Res 28 (2022) 16–24. https://doi.org/10.4258/hir.2022.28.1.16.

[27] Y. Li, R.M. Wehbe, F.S. Ahmad, H. Wang, Y. Luo, A comparative study of pretrained language models for long clinical text, Journal of the American Medical Informatics Association 30 (2023) 340–347. https://doi.org/10.1093/jamia/ocac225.

[28] X. Zhang, Y. Zhang, Q. Zhang, Y. Ren, T. Qiu, J. Ma, Q. Sun, Extracting comprehensive clinical information for breast cancer using deep learning methods, International Journal of Medical Informatics 132 (2019) 103985. https://doi.org/10.1016/j.ijmedinf.2019.103985.

[29] M. Li, F. Liu, J. Zhu, R. Zhang, Y. Qin, D. Gao, Model-based clinical note entity recognition for rheumatoid arthritis using bidirectional encoder representation from transformers, Quant Imaging Med Surg 12 (2022) 184–195. https://doi.org/10.21037/qims-21-90.

[30] Z. Yu, X. Yang, G.L. Sweeting, Y. Ma, S.E. Stolte, R. Fang, Y. Wu, Identify diabetic retinopathy-related clinical concepts and their attributes using transformer-based natural language processing methods, BMC Medical Informatics and Decision Making 22 (2022) 255. https://doi.org/10.1186/s12911-022-01996-2.

[31] Z. Yu, X. Yang, C. Dang, S. Wu, P. Adekkanattu, J. Pathak, T.J. George, W.R. Hogan, Y. Guo, J. Bian, Y. Wu, A Study of Social and Behavioral Determinants of Health in Lung Cancer Patients Using Transformers-based Natural Language Processing Models, AMIA Annu Symp Proc 2021 (2022) 1225–1233.

[32] Guttmacher Alan E., Collins Francis S., Carmona Richard H., The Family History — More Important Than Ever, New England Journal of Medicine 351 (2004) 2333–2336. https://doi.org/10.1056/NEJMsb042979.

[33] R. Schleifer, J. Vannatta, The Logic of Diagnosis: Peirce, Literary Narrative, and the History of Present Illness, The Journal of Medicine and Philosophy: A Forum for Bioethics and Philosophy of Medicine 31 (2006) 363–384. https://doi.org/10.1080/03605310600860809.

[34] Behforouz Heidi L., Drain Paul K., Rhatigan Joseph J., Rethinking the Social History, New England Journal of Medicine 371 (2014) 1277–1279. https://doi.org/10.1056/NEJMp1404846.

[35] J.C. Muhrer, The importance of the history and physical in diagnosis, The Nurse Practitioner 39 (2014) 30. https://doi.org/10.1097/01.NPR.0000444648.20444.e6.

[36] L.A. Calder, G. Mastoras, M. Rahimpour, B. Sohmer, B. Weitzman, A.A. Cwinn, T. Hobin, A. Parush, Team communication patterns in emergency resuscitation: a mixed methods qualitative analysis, International Journal of Emergency Medicine 10 (2017) 24. https://doi.org/10.1186/s12245-017-0149-4.

[37] S.W.H. Lee, D. Thomas, S. Zachariah, J.C. Cooper, Chapter 6 - Communication Skills and Patient History Interview, in: D. Thomas (Ed.), Clinical Pharmacy Education, Practice and Research, Elsevier, 2019: pp. 79–89. https://doi.org/10.1016/B978-0-12-814276-9.00006-4.

[38] CPT® Evaluation and Management, American Medical Association (2023). https://www.ama-assn.org/practice-management/cpt/cpt-evaluation-and-management (accessed June 4, 2024).

[39] Centers for Medicare & Medicaid Services (CMS), 1997 Documentation Guidelines for Evaluation and Management Services | U.S. Dept. of Health & Human Services Guidance Portal, (n.d.). https://www.hhs.gov/guidance/document/1997-documentation-guidelines-evaluation-and-management-services (accessed June 5, 2024).

[40] Y. Wang, L. Wang, M. Rastegar-Mojarad, S. Moon, F. Shen, N. Afzal, S. Liu, Y. Zeng, S. Mehrabi, S. Sohn, H. Liu, Clinical information extraction applications: A literature review, J Biomed Inform 77 (2018) 34–49. https://doi.org/10.1016/j.jbi.2017.11.011.

[41] Hello GPT-4o, (n.d.). https://openai.com/index/hello-gpt-4o/ (accessed October 5, 2024).

[42] Y. Hu, Q. Chen, J. Du, X. Peng, V.K. Keloth, X. Zuo, Y. Zhou, Z. Li, X. Jiang, Z. Lu, K. Roberts, H. Xu, Improving large language models for clinical named entity recognition via prompt engineering, Journal of the American Medical Informatics Association 31 (2024) 1812–1820. https://doi.org/10.1093/jamia/ocad259.

[43] E. Alsentzer, J. Murphy, W. Boag, W.-H. Weng, D. Jindi, T. Naumann, M. McDermott, Publicly Available Clinical BERT Embeddings, in: A. Rumshisky, K. Roberts, S. Bethard, T. Naumann (Eds.), Proceedings of the 2nd Clinical

Natural Language Processing Workshop, Association for Computational Linguistics, Minneapolis, Minnesota, USA, 2019: pp. 72–78. https://doi.org/10.18653/v1/W19-1909.

[44] GuYu, TinnRobert, ChengHao, LucasMichael, UsuyamaNaoto, LiuXiaodong, NaumannTristan, GaoJianfeng, PoonHoifung, Domain-Specific Language Model Pretraining for Biomedical Natural Language Processing, ACM Transactions on Computing for Healthcare (HEALTH) (2021). https://doi.org/10.1145/3458754.

[45] R. Tinn, H. Cheng, Y. Gu, N. Usuyama, X. Liu, T. Naumann, J. Gao, H. Poon, Fine-tuning large neural language models for biomedical natural language processing, Patterns (N Y) 4 (2023) 100729. https://doi.org/10.1016/j.patter.2023.100729.

[46] H.-C. Shin, Y. Zhang, E. Bakhturina, R. Puri, M. Patwary, M. Shoeybi, R. Mani, BioMegatron: Larger Biomedical Domain Language Model, in: B. Webber, T. Cohn, Y. He, Y. Liu (Eds.), Proceedings of the 2020 Conference on Empirical Methods in Natural Language Processing (EMNLP), Association for Computational Linguistics, Online, 2020: pp. 4700–4706. https://doi.org/10.18653/v1/2020.emnlp-main.379.

[47] X. Yang, A. Chen, N. PourNejatian, H.C. Shin, K.E. Smith, C. Parisien, C. Compas, C. Martin, A.B. Costa, M.G. Flores, Y. Zhang, T. Magoc, C.A. Harle, G. Lipori, D.A. Mitchell, W.R. Hogan, E.A. Shenkman, J. Bian, Y. Wu, A large language model for electronic health records, Npj Digit. Med. 5 (2022) 1–9. https://doi.org/10.1038/s41746-022-00742-2.

[48] C. Peng, X. Yang, A. Chen, K.E. Smith, N. PourNejatian, A.B. Costa, C. Martin, M.G. Flores, Y. Zhang, T. Magoc, G. Lipori, D.A. Mitchell, N.S. Ospina, M.M. Ahmed, W.R. Hogan, E.A. Shenkman, Y. Guo, J. Bian, Y. Wu, A study of generative large language model for medical research and healthcare, Npj Digit. Med. 6 (2023) 1–10. https://doi.org/10.1038/s41746-023-00958-w.

[49] J. Devlin, M.-W. Chang, K. Lee, K. Toutanova, BERT: Pre-training of Deep Bidirectional Transformers for Language Understanding, (2019). https://doi.org/10.48550/arXiv.1810.04805.

[50] E. Soysal, J. Wang, M. Jiang, Y. Wu, S. Pakhomov, H. Liu, H. Xu, CLAMP – a toolkit for efficiently building customized clinical natural language processing pipelines, J Am Med Inform Assoc 25 (2017) 331–336. https://doi.org/10.1093/jamia/ocx132.

[51] T. Thieu, J.C. Maldonado, P.-S. Ho, M. Ding, A. Marr, D. Brandt, D. Newman-Griffis, A. Zirikly, L. Chan, E. Rasch, A comprehensive study of mobility functioning information in clinical notes: Entity hierarchy, corpus annotation, and sequence labeling, International Journal of Medical Informatics 147 (2021) 104351. https://doi.org/10.1016/j.ijmedinf.2020.104351.

[52] N. Chinchor, B. Sundheim, MUC-5 Evaluation Metrics, in: Fifth Message Understanding Conference (MUC-5): Proceedings of a Conference Held in Baltimore, Maryland, August 25-27, 1993, 1993. https://aclanthology.org/M93-1007 (accessed June 5, 2024).

[53] J. Gillis, The History of the Patient History since 1850, Bulletin of the History of Medicine 80 (2006) 490–512.

[54] Y. Zhang, P. Yu, J. Shen, The benefits of introducing electronic health records in residential aged care facilities: A multiple case study, International Journal of Medical Informatics 81 (2012) 690–704. https://doi.org/10.1016/j.ijmedinf.2012.05.013.

[55] A. Noblin, K. Cortelyou-Ward, J. Cantiello, T. Breyer, L. Oliveira, M. Dangiolo, M. Cannarozzi, T. Yeung, S. Berman, EHR Implementation in a New Clinic: A Case Study of Clinician Perceptions, J Med Syst 37 (2013) 9955. https://doi.org/10.1007/s10916-013-9955-2.

[56] M. Tayefi, P. Ngo, T. Chomutare, H. Dalianis, E. Salvi, A. Budrionis, F. Godtliebsen, Challenges and opportunities beyond structured data in analysis of electronic health records, WIREs Computational Statistics 13 (2021) e1549. https://doi.org/10.1002/wics.1549.

[57] M. Agrawal, S. Hegselmann, H. Lang, Y. Kim, D. Sontag, Large Language Models are Few-Shot Clinical Information Extractors, (2022). http://arxiv.org/abs/2205.12689 (accessed June 6, 2024).

[58] C. Wang, S. Liu, H. Yang, J. Guo, Y. Wu, J. Liu, Ethical Considerations of Using ChatGPT in Health Care, Journal of Medical Internet Research 25 (2023) e48009. https://doi.org/10.2196/48009.

[59] P.P. Ray, Timely need for navigating the potential and downsides of LLMs in healthcare and biomedicine, Briefings in Bioinformatics 25 (2024) bbae214. https://doi.org/10.1093/bib/bbae214.

[60] Meta Llama, Meta Llama (n.d.). https://llama.meta.com/ (accessed June 6, 2024).

[61] J.A. Marron, Implementing the Health Insurance Portability and Accountability Act (HIPAA) security rule: a cybersecurity resource guide, National Institute of Standards and Technology (U.S.), Gaithersburg, MD, 2024. https://doi.org/10.6028/NIST.SP.800-66r2.

[62] V.V. Thaker, F. Lee, C.J. Bottino, C.L. Perry, I.A. Holm, J.N. Hirschhorn, S.K. Osganian, Impact of an Electronic Template on Documentation of Obesity in a Primary Care Clinic, Clin Pediatr (Phila) 55 (2016) 1152–1159. https://doi.org/10.1177/0009922815621331.

[63] E.Y. Rieger, I.J. Anderson, V.G. Press, M.X. Cui, V.M. Arora, B.C. Williams, J.W. Tang, Implementation of a Biopsychosocial History and Physical Exam Template in the Electronic Health Record: Mixed Methods Study, JMIR Med Educ 9 (2023) e42364. https://doi.org/10.2196/42364.

[64] M. Neumann, D. King, I. Beltagy, W. Ammar, ScispaCy: Fast and Robust Models for Biomedical Natural Language Processing, in: D. Demner-Fushman, K.B. Cohen, S. Ananiadou, J. Tsujii (Eds.), Proceedings of the 18th BioNLP Workshop and Shared Task, Association for Computational Linguistics, Florence, Italy, 2019: pp. 319–327. https://doi.org/10.18653/v1/W19-5034.

[65] E. Kıcıman, R. Ness, A. Sharma, C. Tan, Causal Reasoning and Large Language Models: Opening a New Frontier for Causality, (2023). https://doi.org/10.48550/arXiv.2305.00050.

[66] J. Zhang, M. Wibert, H. Zhou, X. Peng, Q. Chen, V.K. Keloth, Y. Hu, R. Zhang, H. Xu, K. Raja, A Study of Biomedical Relation Extraction Using GPT Models, AMIA Jt Summits Transl Sci Proc 2024 (2024) 391–400.

[67] C. Elkan, The foundations of cost-sensitive learning, in: International Joint Conference on Artificial Intelligence, Lawrence Erlbaum Associates Ltd, 2001: pp. 973–978. http://cseweb.ucsd.edu/~elkan/rescale.pdf (accessed July 26, 2024).

# Supplementary Material-1: Examples of Historical Entities

Table S1: Definitions of CC/HPI/PFSH concepts and associated examples according to the 1997 E/M Guidelines.

| History | Concepts | Definitions | Examples |
|---|---|---|---|
| CC | | A brief statement on the reason for a medical encounter, usually stated in the patient's own words. | Migraine |
| HPI | Location | Anatomical location of the problem | Right-side of my head |
| | Quality | The nature of the problem | Constant, aching pain |
| | Severity | A degree or measurement of how bad the problem is | 5 on a scale of 1-10 |
| | Duration | How long the problem has been existing | It started two days ago |
| | Timing | The temporal pattern of the problem, such as when and how frequently it occurs. | One or two episodes of pains per day, all happened in afternoons |
| | Context | The patient's activities, environmental factors, and/or circumstances surrounding the problem. | I stayed up late every night to catch up a deadline last week |
| | Modifying Factors | Factors that make the problem better or worse | Better when heat is applied |
| | Associated Signs or Symptoms | Symptoms or signs that accompany the problem | Vomiting and fatigue |
| PFSH | Past Medical History | The patient's past experiences with illness, treatments, and operations | Pre-diabetes and hypertension |
| | Family History | Medical events in the patient's family, such as diseases which are hereditary or place the patient at risk | Father has migraine as well |
| | Social History | An age-appropriate review of the patient's social activities | Nonsmoker, drinks occasionally |

Selected examples showing the polysemous nature of historical concepts. Entities are highlighted with red, bold font.

- HPI-Modifying Factors can include multiple sub-concepts like Medications, Environment Factors, and Actions:

    o Medication: "The patient returns to our office today because of continued problems with her headaches. … She takes **Motrin** 400 mg b.i.d., which helped." in sample_1128.txt.

    o Environmental Factor: "Her husband has been **hauling corn and this seems to aggravate things**." in sample_343.txt.

    o Action: "She has little bit of paraesthesias in the left toe as well and seem to involve all the toes of the right foot. They are not worse with walking. It seems to be **worse when she is in bed**." in sample_365.txt.

- HPI-Associated Signs and Symptoms can include sub-concepts like Problems and Signs, among which Signs may not be immediately apparent.
  - Problem: "When questioned further, she described **shortness of breath** primarily with ambulation." in sample_398.txt.
  - Sign: "… the patient continues to **lose weight**." in sample_378.txt.
- Past Medical History can include multiple sub-concepts like Conditions and Treatments:
  - Condition: "PAST MEDICAL HISTORY: "…, **diabetics** with a bad family history for **cardiovascular disease** such as this patient does have, …" in sample_1568.txt.
  - Treatments: "PAST MEDICAL HISTORY: … 5. **Removal of the melanoma from the right thigh** in 1984. …" in sample_380.txt.

## Selected examples of entities that primarily consist of non-medical vocabulary.

- HPI-Context example: "This is a 32-year-old male who **had a piece of glass fall on to his right foot** today." in sample_2747.txt.
- Social History example: "SOCIAL HISTORY: … He has what sounds like a **data entry computer job**." in sample_1152.txt.
- HPI-Quality example: "This is a one plus-month-old female with respiratory symptoms … This involved cough, … The coughing persisted and **worsened**." in sample_1956.txt.
- HPI-Timing example: "This is an 18-month-old white male here with his mother for complaint of intermittent fever … Mother states that his temperature usually **elevates at night**." in sample_439.txt

## Selected examples of entities that did not occur in the designated headed sections.

- Example of an HPI-Duration entity in the CC section: "CHIEF COMPLAINT: Cough and abdominal pain for **two days**." in sample_930.txt.
- Example of an HPI-Location entity in the CC section: "CHIEF COMPLAINT: Pressure decubitus, **right hip**. HISTORY OF PRESENT ILLNESS: This is a 30-year-old female patient presenting with the above chief complaint." in sample_687.txt.
- Examples of Past Medical History entities in the HPI/Subjective section:
  - "SUBJECTIVE: This is a 78-year-old male who recently had his **right knee replaced** and also **back surgery** about a year and a half ago." in sample_70.txt.
  - "BRIEF HISTORY OF PRESENT ILLNESS: … The patient has history of **hypercholesterolemia**, …" in sample_96.txt.
  - "HISTORY OF PRESENT ILLNESS: This is the initial clinic visit for a 29-year-old man … He has **no upper extremity**." in sample_223.txt.
- Examples of CC entities in the HPI/Subjective section:

- "HISTORY OF PRESENT ILLNESS: This is the initial clinic visit for a 29-year-old man who is seen for new onset of **right shoulder pain**." in sample_223.txt.
- "HISTORY OF PRESENT ILLNESS: This is the initial clinic visit for a 41-year-old worker who is seen for a **foreign body to his left eye**." in sample_225.txt.

- Examples of Social History entities in the HPI/Subjective section:
  - "CHIEF COMPLAINT: Blood-borne pathogen exposure. HISTORY OF PRESENT ILLNESS: The patient is a 54-year-old right-handed male who **works as a phlebotomist and respiratory therapist at Hospital**." in sample_226.txt.
  - "HISTORY OF PRESENT ILLNESS: The patient is a 55-year-old Caucasian female … The patient is a **nonsmoker**." in sample_380.txt.

- Examples of Family History entities in the HPI/Subjective section:
  - "SUBJECTIVE: The patient is a 7-year-old male who comes in today with a three-day history of emesis and a four-day history of diarrhea. Apparently, **his brother had similar symptoms**." in sample_388.txt.
  - "HISTORY OF PRESENT ILLNESS: This 60-year-old white male is referred to us … There is a history of **gallstone pancreatitis** in the patient's sister." in sample_2780.txt.

# Supplementary Material-2: Model Development and Specifications

## Prompt for GPT-4o

### Task

Your task is to generate an HTML version of an input text, marking up specific entities related to healthcare which are in doctor's note or clinical note.

The entities to be identified are: 'cc', 'hpi.location', 'hpi.quality', 'hpi.severity', 'hpi.duration', 'hpi.timing', 'hpi.context', 'hpi.modifyingFactors', 'hpi.assocSignsAndSymptoms', 'pastHistory','familyHistory', 'socialHistory'.

If a sentence has negation words, entities might not need to be identified.

Use HTML <span> tags to highlight these entities. Each <span> should have a class attribute indicating the type of the entity.

### Entity Markup Guide

Use <span class="cc"> to denote a chief complain entity in the clinical note

Use <span class="hpi.location"> to denote an entity related to the location of a symptom or condition in the history of present illness.

Use <span class="hpi.quality"> to denote an entity related to the quality or character of a symptom in the history of present illness.

Use <span class="hpi.severity"> to denote an entity related to the severity of a symptom or condition in the history of present illness.

Use <span class="hpi.duration"> to denote an entity related to how long a symptom or condition has been present in the history of present illness.

Use <span class="hpi.timing"> to denote an entity related to the timing or frequency of a symptom in the history of present illness.

Use <span class="hpi.context"> to denote an entity related to the context or circumstances surrounding a symptom or condition in the history of present illness.

Use <span class="hpi.modifyingFactors"> to denote an entity related to factors that make a symptom better or worse in the history of present illness.

Use <span class="hpi.assocSignsAndSymptoms"> to denote an entity related to associated signs and symptoms present with the condition.

Use <span class="pastHistory"> to denote an entity related to the patient's past medical history.

Use <span class="familyHistory"> to denote an entity related to the patient's family history.

Use <span class="socialHistory"> to denote an entity related to the patient's social history, such as lifestyle or habits.

Leave the text as it is if no such entities are found.

## CLLMs Evaluated

- **Bio+Discharge Summary BERT & Bio+Clinical BERT**. Both the models were initialized with BioBERT [1], specifically, starting with BioBERT's architecture and parameters, which include 12 transformer layers and approximately 110 million parameters. Afterward, Bio+Discharge Summary BERT was fine-tuned on the discharge summaries in the MIMIC-III EHR repository [2], while Bio+Clinical BERT underwent fine-tuning with all clinical notes from MIMIC-III. The fine-tuning process involved further adjusting the model's parameters based on the clinical corpora provided, enabling the models to specialize in understanding the semantics within the given categories of clinical text [3].

- **PubMedBERT & PubMedBERT Large**. The PubMedBERT model was created by training the BERT$_{BASE}$ architecture [4], which consists of 12 layers and 110 million parameters, on the abstracts from PubMed and the full-text articles from PubMed Central. The development of PubMedBERT Large was based on the BERT$_{LARGE}$ architecture with 24 layers and 340 million parameters, but utilizing only the abstracts from PubMed [5,6].

- **BioMegatron**. The BioMegatron model was initiated with Megatron-LM [7], then fine-tuned using the abstracts from PubMed and the full-text articles from PubMedCentral. While different sizes of BioMegatron are available [8], we employed the one with 24 layers and 345 million parameters in this study.

- **GatorTron & GatorTronS**. Trained on a variety of copra including PubMed, WikiText, MIMIC-III clinical notes, and de-identified clinical notes from the University of Florida Health System (for GatorTron) or synthetic clinical words generated by GPT-3 model (for GatorTronS), GatorTron and GatorTronS stand out as two of the most extensively trained cLLMs [9,10]. The GatorTron and GatorTronS models, we used in this study, each had 24 layers and 345 million parameters.

## The Two Fine-Tuning Approaches

- **Basic Fine-Tuning**: It is achieved by building a fully connected output layer on top of a cLLM [4]. Specifically, the process begins with feeding textual data from clinical notes into the cLLM in batches, with each batch containing eight sentences. Given an $m$-token batch, the cLLM transforms the tokens into $m$ vectors of $n$-dimensions ($n$ depends on the specific cLLM used), denoted as $\{V_1, V_2, \ldots, V_m\}$. Each vector represents the embedding and positional encoding of a token. These vectors are then passed to the output layer for classification. Given that 12 medical history entities (MHEs) resulted in a total of 25 *"B-<MHE>", "I-<MHE>",* and *"O"* tags, the output layer comprises 25 neurons corresponding to the 25 BIO tags. Each neuron contains $n$ weights. The classification will select the BIO tag corresponding to the neuron that has the highest dot summation between the embedding vector and the weights. Thus, the output layer has $n \times 25$ weights, denoted as $W \in \mathbb{R}^{n \times 25}$, which need to be trained for optimal classification. Let $\{Y_1, Y_2, \ldots, Y_m\}$ be the BIO tags of the $m$-token batch, the training process is directed towards determining the optimal $W$ that minimizes the following cross-entropy loss function.

$$\min_{W \in \mathbb{R}^{n \times 25}} L = -\sum_{i=1}^{m} CrossEntropy_i(V_i, Y_i; W)$$

Furthermore, we applied the dropout technique to mitigate the potential overfitting [11]. In our implementation, dropout was used to randomly deactivates 10% elements of the embedding vector output by the cLLM, by setting them to 0.

- **Fine-Tuning With Pre-Identified BMEs**: The BMEs pre-identified by CLAMP can be categorized into two groups:

    o *Group 1*: "problem," "test," "treatment," and "drug," and

    o *Group 2*: "body location," "severity," and "temporal."

    These two groups of entities can overlap. For example, the term "chest pain" is classified as a "problem," whereas "chest" in this context refers to a "body location." To explore the potential of the preliminary BME information in enhancing the recognition of sophisticated CC/HPI/PFSH entities, we incorporated it into the cLLM fine-tuning as follows:

    1) *New Tag Creation*: We added new BIO tags, specifically "B-<BME>," "I-<BME>," and "O," to each token based on the BME identification provided by CLAMP. Given the seven types of BMEs, this resulted in a total of 15 BIO tags. Note that, due to the overlapping of the two groups of CLAMP BMEs, it is possible that a token can have multiple tags. For instance, the "chest" in term "chest pain" can have both "B-problem" and "B-Body Location" tags.

    2) *One-Hot Encoding*: We applied one-hot encoding to convert each token's BME-related BIO tag into a 14-dimensional vector. A value of 1 in a vector element indicates a specific "B-<BME>" or "I-<BME>" tag, while all 0s correspond to the "O" tag.

    3) *Concatenation and Fine-Tuning*: The one-hot encoded vectors representing BME information were concatenated with the cLLM embeddings. Each concatenated vector was then fine-tuned towards the 25 historical entity-related BIO tags in a manner similar to the basic fine-tuning process described earlier.

All cLLMs were implemented using PyTorch and trained for 40 epochs. The code is available at https://github.com/isallexist/Patient-History-NER.

## References

[1] J. Lee, W. Yoon, S. Kim, D. Kim, S. Kim, C.H. So, J. Kang, BioBERT: a pre-trained biomedical language representation model for biomedical text mining, Bioinformatics 36 (2020) 1234–1240. https://doi.org/10.1093/bioinformatics/btz682.

[2] A.E.W. Johnson, T.J. Pollard, L. Shen, L.H. Lehman, M. Feng, M. Ghassemi, B. Moody, P. Szolovits, L. Anthony Celi, R.G. Mark, MIMIC-III, a freely accessible critical care database, Scientific Data 3 (2016) 160035. https://doi.org/10.1038/sdata.2016.35.

[3] E. Alsentzer, J. Murphy, W. Boag, W.-H. Weng, D. Jindi, T. Naumann, M. McDermott, Publicly Available Clinical BERT Embeddings, in: A. Rumshisky, K. Roberts, S. Bethard, T. Naumann (Eds.), Proceedings of the 2nd Clinical Natural Language Processing Workshop, Association for Computational Linguistics, Minneapolis, Minnesota, USA, 2019: pp. 72–78. https://doi.org/10.18653/v1/W19-1909.

[4] J. Devlin, M.-W. Chang, K. Lee, K. Toutanova, BERT: Pre-training of Deep Bidirectional Transformers for Language Understanding, (2019). https://doi.org/10.48550/arXiv.1810.04805.

[5] GuYu, TinnRobert, ChengHao, LucasMichael, UsuyamaNaoto, LiuXiaodong, NaumannTristan, GaoJianfeng, PoonHoifung, Domain-Specific Language Model Pretraining for Biomedical Natural Language Processing, ACM Transactions on Computing for Healthcare (HEALTH) (2021). https://doi.org/10.1145/3458754.

[6] R. Tinn, H. Cheng, Y. Gu, N. Usuyama, X. Liu, T. Naumann, J. Gao, H. Poon, Fine-tuning large neural language models for biomedical natural language processing, Patterns (N Y) 4 (2023) 100729. https://doi.org/10.1016/j.patter.2023.100729.

[7] M. Shoeybi, M. Patwary, R. Puri, P. LeGresley, J. Casper, B. Catanzaro, Megatron-LM: Training Multi-Billion Parameter Language Models Using Model Parallelism, (2020). https://doi.org/10.48550/arXiv.1909.08053.

[8] H.-C. Shin, Y. Zhang, E. Bakhturina, R. Puri, M. Patwary, M. Shoeybi, R. Mani, BioMegatron: Larger Biomedical Domain Language Model, in: B. Webber, T. Cohn, Y. He, Y. Liu (Eds.), Proceedings of the 2020 Conference on Empirical Methods in Natural Language Processing (EMNLP), Association for Computational Linguistics, Online, 2020: pp. 4700–4706. https://doi.org/10.18653/v1/2020.emnlp-main.379.

[9] X. Yang, A. Chen, N. PourNejatian, H.C. Shin, K.E. Smith, C. Parisien, C. Compas, C. Martin, A.B. Costa, M.G. Flores, Y. Zhang, T. Magoc, C.A. Harle, G. Lipori, D.A. Mitchell, W.R. Hogan, E.A. Shenkman, J. Bian, Y. Wu, A large language model for electronic health records, Npj Digit. Med. 5 (2022) 1–9. https://doi.org/10.1038/s41746-022-00742-2.

[10] C. Peng, X. Yang, A. Chen, K.E. Smith, N. PourNejatian, A.B. Costa, C. Martin, M.G. Flores, Y. Zhang, T. Magoc, G. Lipori, D.A. Mitchell, N.S. Ospina, M.M. Ahmed, W.R. Hogan, E.A. Shenkman, Y. Guo, J. Bian, Y. Wu, A study of generative large language model for medical research and healthcare, Npj Digit. Med. 6 (2023) 1–10. https://doi.org/10.1038/s41746-023-00958-w.

[11] N. Srivastava, G. Hinton, A. Krizhevsky, I. Sutskever, R. Salakhutdinov, Dropout: a simple way to prevent neural networks from overfitting, The Journal of Machine Learning Research 15 (2014) 1929–1958.

# Supplementary Material-3: Additional Results

Table S2: Detailed counts and rates of matching results for each CC/HPI/PFSH concept.

| Entity | Count | EM | RM | MM | UD | OD | MMUD | Error | RM+Error |
|---|---|---|---|---|---|---|---|---|---|
| colspan="9" | OpenAI GPT-4o with zero-short learning |||||||||
| CC | 133 | 53 (39.8%) | 27 (20.3%) | 38 (28.6%) | 15 (11.3%) | 306 (230.1%) | 53 (39.8%) | 359 (269.9%) | 386 (290.2%) |
| Location | 48 | 24 (50.0%) | 26 (54.2%) | 3 (6.3%) | 2 (4.2%) | 283 (589.6%) | 5 (10.4%) | 288 (600.0%) | 314 (654.2%) |
| Quality | 53 | 8 (15.1%) | 12 (22.6%) | 9 (17.0%) | 25 (47.2%) | 124 (234.0%) | 34 (64.2%) | 158 (298.1%) | 170 (320.8%) |
| Severity | 34 | 16 (47.1%) | 5 (14.7%) | 5 (14.7%) | 8 (23.5%) | 50 (147.1%) | 13 (38.2%) | 63 (185.3%) | 68 (200.0%) |
| Duration | 66 | 25 (37.9%) | 21 (31.8%) | 5 (7.6%) | 15 (22.7%) | 85 (128.8%) | 20 (30.3%) | 105 (159.1%) | 126 (190.9%) |
| Timing | 37 | 7 (18.9%) | 8 (21.6%) | 10 (27.0%) | 13 (35.1%) | 42 (113.5%) | 23 (62.2%) | 65 (175.7%) | 73 (197.3%) |
| Context | 37 | 1 (2.7%) | 3 (8.1%) | 14 (37.8%) | 19 (51.4%) | 26 (70.3%) | 33 (89.2%) | 59 (159.5%) | 62 (167.6%) |
| M.F. | 81 | 9 (11.1%) | 4 (4.9%) | 10 (12.3%) | 58 (71.6%) | 28 (34.6%) | 68 (84.0%) | 96 (118.5%) | 100 (123.5%) |
| Symptom | 268 | 54 (20.1%) | 17 (6.3%) | 74 (27.6%) | 124 (46.3%) | 268 (100.0%) | 198 (73.9%) | 466 (173.9%) | 483 (180.2%) |
| Past H. | 518 | 120 (23.2%) | 56 (10.8%) | 19 (3.7%) | 323 (62.4%) | 205 (39.6%) | 342 (66.0%) | 547 (105.6%) | 603 (116.4%) |
| Fam. H. | 45 | 23 (51.1%) | 18 (40.0%) | 0 (0.0%) | 5 (11.1%) | 39 (86.7%) | 5 (11.1%) | 44 (97.8%) | 62 (137.8%) |
| Social H. | 129 | 23 (17.8%) | 25 (19.4%) | 4 (3.1%) | 82 (63.6%) | 42 (32.6%) | 86 (66.7%) | 128 (99.2%) | 153 (118.6%) |
| *Total* | 1449 | 363 (25.1%) | 222 (15.3%) | 191 (13.2%) | 689 (47.6%) | 1498 (103.4%) | 880 (60.7%) | 2378 (164.1%) | 2600 (179.4%) |
| colspan="9" | Bio+Discharge Summary BERT |||||||||
| Entity | Count | EM | RM | MM | UD | OD | MMUD | Error | RM+Error |
| CC | 133 | 51 (38.3%) | 14 (10.5%) | 40 (30.1%) | 29 (21.8%) | 16 (12.0%) | 69 (51.9%) | 85 (63.9%) | 99 (74.4%) |
| Location | 48 | 13 (27.1%) | 12 (25.0%) | 9 (18.8%) | 16 (33.3%) | 15 (31.3%) | 25 (52.1%) | 40 (83.3%) | 52 (108.3%) |
| Quality | 53 | 11 (20.8%) | 10 (18.9%) | 9 (17.0%) | 25 (47.2%) | 14 (26.4%) | 34 (64.2%) | 48 (90.6%) | 58 (109.4%) |
| Severity | 34 | 3 (8.8%) | 3 (8.8%) | 6 (17.6%) | 22 (64.7%) | 7 (20.6%) | 28 (82.4%) | 35 (102.9%) | 38 (111.8%) |
| Duration | 66 | 27 (40.9%) | 17 (25.8%) | 0 (0.0%) | 23 (34.8%) | 29 (43.9%) | 23 (34.8%) | 52 (78.8%) | 69 (104.5%) |
| Timing | 37 | 13 (35.1%) | 0 (0.0%) | 11 (29.7%) | 13 (35.1%) | 11 (29.7%) | 24 (64.9%) | 35 (94.6%) | 35 (94.6%) |
| Context | 37 | 1 (2.7%) | 7 (18.9%) | 17 (45.9%) | 12 (32.4%) | 13 (35.1%) | 29 (78.4%) | 42 (113.5%) | 49 (132.4%) |
| M.F. | 81 | 34 (42.0%) | 3 (3.7%) | 11 (13.6%) | 33 (40.7%) | 36 (44.4%) | 44 (54.3%) | 80 (98.8%) | 83 (102.5%) |
| Symptom | 268 | 74 (27.6%) | 49 (18.3%) | 27 (10.1%) | 122 (45.5%) | 143 (53.4%) | 149 (55.6%) | 292 (109.0%) | 341 (127.2%) |

| Entity | Count | EM | RM | MM | UD | OD | MMUD | Error | RM+Error |
|---|---|---|---|---|---|---|---|---|---|
| Past H. | 518 | 351 (67.8%) | 38 (7.3%) | 19 (3.7%) | 113 (21.8%) | 152 (29.3%) | 132 (25.5%) | 284 (54.8%) | 322 (62.2%) |
| Fam. H. | 45 | 34 (75.6%) | 5 (11.1%) | 2 (4.4%) | 5 (11.1%) | 4 (8.9%) | 7 (15.6%) | 11 (24.4%) | 16 (35.6%) |
| Social H. | 129 | 60 (46.5%) | 54 (41.9%) | 5 (3.9%) | 20 (15.5%) | 22 (17.1%) | 25 (19.4%) | 47 (36.4%) | 101 (78.3%) |
| *Total* | 1449 | 672 (46.4%) | 212 (14.6%) | 156 (10.8%) | 433 (29.9%) | 462 (31.9%) | 589 (40.6%) | 1051 (72.5%) | 1263 (87.2%) |
| Bio+Discharge Summary BERT + CLAMP | | | | | | | | | |
| Entity | Count | EM | RM | MM | UD | OD | MMUD | Error | RM+Error |
| CC | 133 | 66 (49.6%) | 23 (17.3%) | 30 (22.6%) | 18 (13.5%) | 24 (18.0%) | 48 (36.1%) | 72 (54.1%) | 95 (71.4%) |
| Location | 48 | 11 (22.9%) | 12 (25.0%) | 9 (18.8%) | 19 (39.6%) | 11 (22.9%) | 28 (58.3%) | 39 (81.3%) | 51 (106.3%) |
| Quality | 53 | 11 (20.8%) | 10 (18.9%) | 7 (13.2%) | 26 (49.1%) | 14 (26.4%) | 33 (62.3%) | 47 (88.7%) | 57 (107.5%) |
| Severity | 34 | 3 (8.8%) | 2 (5.9%) | 11 (32.4%) | 18 (52.9%) | 7 (20.6%) | 29 (85.3%) | 36 (105.9%) | 38 (111.8%) |
| Duration | 66 | 25 (37.9%) | 15 (22.7%) | 0 (0.0%) | 27 (40.9%) | 26 (39.4%) | 27 (40.9%) | 53 (80.3%) | 68 (103.0%) |
| Timing | 37 | 11 (29.7%) | 0 (0.0%) | 14 (37.8%) | 12 (32.4%) | 12 (32.4%) | 26 (70.3%) | 38 (102.7%) | 38 (102.7%) |
| Context | 37 | 2 (5.4%) | 9 (24.3%) | 18 (48.6%) | 9 (24.3%) | 10 (27.0%) | 27 (73.0%) | 37 (100.0%) | 46 (124.3%) |
| M.F. | 81 | 31 (38.3%) | 3 (3.7%) | 14 (17.3%) | 33 (40.7%) | 29 (35.8%) | 47 (58.0%) | 76 (93.8%) | 79 (97.5%) |
| Symptom | 268 | 71 (26.5%) | 27 (10.1%) | 37 (13.8%) | 136 (50.7%) | 130 (48.5%) | 173 (64.6%) | 303 (113.1%) | 330 (123.1%) |
| Past H. | 518 | 338 (65.3%) | 48 (9.3%) | 15 (2.9%) | 125 (24.1%) | 158 (30.5%) | 140 (27.0%) | 298 (57.5%) | 346 (66.8%) |
| Fam. H. | 45 | 35 (77.8%) | 5 (11.1%) | 1 (2.2%) | 5 (11.1%) | 3 (6.7%) | 6 (13.3%) | 9 (20.0%) | 14 (31.1%) |
| Social H. | 129 | 61 (47.3%) | 52 (40.3%) | 4 (3.1%) | 20 (15.5%) | 20 (15.5%) | 24 (18.6%) | 44 (34.1%) | 96 (74.4%) |
| *Total* | 1449 | 665 (45.9%) | 206 (14.2%) | 160 (11.0%) | 448 (30.9%) | 444 (30.6%) | 608 (42.0%) | 1052 (72.6%) | 1258 (86.8%) |
| Bio+Clinical BERT | | | | | | | | | |
| Entity | Count | EM | RM | MM | UD | OD | MMUD | Error | RM+Error |
| CC | 133 | 43 (32.3%) | 8 (6.0%) | 44 (33.1%) | 38 (28.6%) | 16 (12.0%) | 82 (61.7%) | 98 (73.7%) | 106 (79.7%) |
| Location | 48 | 9 (18.8%) | 8 (16.7%) | 8 (16.7%) | 25 (52.1%) | 17 (35.4%) | 33 (68.8%) | 50 (104.2%) | 58 (120.8%) |
| Quality | 53 | 17 (32.1%) | 7 (13.2%) | 7 (13.2%) | 23 (43.4%) | 17 (32.1%) | 30 (56.6%) | 47 (88.7%) | 54 (101.9%) |
| Severity | 34 | 5 (14.7%) | 1 (2.9%) | 10 (29.4%) | 18 (52.9%) | 10 (29.4%) | 28 (82.4%) | 38 (111.8%) | 39 (114.7%) |
| Duration | 66 | 31 (47.0%) | 16 (24.2%) | 2 (3.0%) | 20 (30.3%) | 32 (48.5%) | 22 (33.3%) | 54 (81.8%) | 70 (106.1%) |
| Timing | 37 | 15 (40.5%) | 1 (2.7%) | 9 (24.3%) | 12 (32.4%) | 15 (40.5%) | 21 (56.8%) | 36 (97.3%) | 37 (100.0%) |
| Context | 37 | 0 (0.0%) | 10 (27.0%) | 19 (51.4%) | 10 (27.0%) | 12 (32.4%) | 29 (78.4%) | 41 (110.8%) | 51 (137.8%) |
| M.F. | 81 | 39 (48.1%) | 5 (6.2%) | 2 (2.5%) | 35 (43.2%) | 37 (45.7%) | 37 (45.7%) | 74 (91.4%) | 79 (97.5%) |
| Symptom | 268 | 73 (27.2%) | 55 (20.5%) | 29 (10.8%) | 118 (44.0%) | 164 (61.2%) | 147 (54.9%) | 311 (116.0%) | 366 (136.6%) |
| Past H. | 518 | 346 (66.8%) | 40 (7.7%) | 27 (5.2%) | 109 (21.0%) | 137 (26.4%) | 136 (26.3%) | 273 (52.7%) | 313 (60.4%) |

| Entity | Count | EM | RM | MM | UD | OD | MMUD | Error | RM+Error |
|---|---|---|---|---|---|---|---|---|---|
| Fam. H. | 45 | 25 (55.6%) | 4 (8.9%) | 11 (24.4%) | 5 (11.1%) | 6 (13.3%) | 16 (35.6%) | 22 (48.9%) | 26 (57.8%) |
| Social H. | 129 | 58 (45.0%) | 53 (41.1%) | 8 (6.2%) | 21 (16.3%) | 19 (14.7%) | 29 (22.5%) | 48 (37.2%) | 101 (78.3%) |
| *Total* | 1449 | 661 (45.6%) | 208 (14.4%) | 176 (12.1%) | 434 (30.0%) | 482 (33.3%) | 610 (42.1%) | 1092 (75.4%) | 1300 (89.7%) |
| Bio+Clinical BERT + CLAMP | | | | | | | | | |
| Entity | Count | EM | RM | MM | UD | OD | MMUD | Error | RM+Error |
| CC | 133 | 56 (42.1%) | 20 (15.0%) | 36 (27.1%) | 24 (18.0%) | 27 (20.3%) | 60 (45.1%) | 87 (65.4%) | 107 (80.5%) |
| Location | 48 | 10 (20.8%) | 9 (18.8%) | 13 (27.1%) | 19 (39.6%) | 12 (25.0%) | 32 (66.7%) | 44 (91.7%) | 53 (110.4%) |
| Quality | 53 | 11 (20.8%) | 12 (22.6%) | 8 (15.1%) | 26 (49.1%) | 22 (41.5%) | 34 (64.2%) | 56 (105.7%) | 68 (128.3%) |
| Severity | 34 | 3 (8.8%) | 3 (8.8%) | 8 (23.5%) | 20 (58.8%) | 9 (26.5%) | 28 (82.4%) | 37 (108.8%) | 40 (117.6%) |
| Duration | 66 | 30 (45.5%) | 13 (19.7%) | 1 (1.5%) | 22 (33.3%) | 30 (45.5%) | 23 (34.8%) | 53 (80.3%) | 66 (100.0%) |
| Timing | 37 | 14 (37.8%) | 0 (0.0%) | 10 (27.0%) | 13 (35.1%) | 13 (35.1%) | 23 (62.2%) | 36 (97.3%) | 36 (97.3%) |
| Context | 37 | 1 (2.7%) | 5 (13.5%) | 18 (48.6%) | 14 (37.8%) | 10 (27.0%) | 32 (86.5%) | 42 (113.5%) | 47 (127.0%) |
| M.F. | 81 | 34 (42.0%) | 4 (4.9%) | 10 (12.3%) | 35 (43.2%) | 43 (53.1%) | 45 (55.6%) | 88 (108.6%) | 92 (113.6%) |
| Symptom | 268 | 60 (22.4%) | 43 (16.0%) | 45 (16.8%) | 126 (47.0%) | 151 (56.3%) | 171 (63.8%) | 322 (120.1%) | 365 (136.2%) |
| Past H. | 518 | 350 (67.6%) | 48 (9.3%) | 21 (4.1%) | 105 (20.3%) | 173 (33.4%) | 126 (24.3%) | 299 (57.7%) | 347 (67.0%) |
| Fam. H. | 45 | 31 (68.9%) | 1 (2.2%) | 8 (17.8%) | 5 (11.1%) | 6 (13.3%) | 13 (28.9%) | 19 (42.2%) | 20 (44.4%) |
| Social H. | 129 | 61 (47.3%) | 59 (45.7%) | 7 (5.4%) | 19 (14.7%) | 29 (22.5%) | 26 (20.2%) | 55 (42.6%) | 114 (88.4%) |
| *Total* | 1449 | 661 (45.6%) | 217 (15.0%) | 185 (12.8%) | 428 (29.5%) | 525 (36.2%) | 613 (42.3%) | 1138 (78.5%) | 1355 (93.5%) |
| GatorTron | | | | | | | | | |
| Entity | Count | EM | RM | MM | UD | OD | MMUD | Error | RM+Error |
| CC | 133 | 69 (51.9%) | 11 (8.3%) | 38 (28.6%) | 16 (12.0%) | 14 (10.5%) | 54 (40.6%) | 68 (51.1%) | 79 (59.4%) |
| Location | 48 | 14 (29.2%) | 18 (37.5%) | 7 (14.6%) | 16 (33.3%) | 10 (20.8%) | 23 (47.9%) | 33 (68.8%) | 51 (106.3%) |
| Quality | 53 | 17 (32.1%) | 14 (26.4%) | 5 (9.4%) | 19 (35.8%) | 14 (26.4%) | 24 (45.3%) | 38 (71.7%) | 52 (98.1%) |
| Severity | 34 | 6 (17.6%) | 3 (8.8%) | 9 (26.5%) | 16 (47.1%) | 9 (26.5%) | 25 (73.5%) | 34 (100.0%) | 37 (108.8%) |
| Duration | 66 | 40 (60.6%) | 16 (24.2%) | 1 (1.5%) | 12 (18.2%) | 24 (36.4%) | 13 (19.7%) | 37 (56.1%) | 53 (80.3%) |
| Timing | 37 | 16 (43.2%) | 6 (16.2%) | 9 (24.3%) | 8 (21.6%) | 12 (32.4%) | 17 (45.9%) | *29 (78.4%)* | 35 (94.6%) |
| Context | 37 | 5 (13.5%) | 10 (27.0%) | 13 (35.1%) | 9 (24.3%) | 10 (27.0%) | 22 (59.5%) | *32 (86.5%)* | 42 (113.5%) |
| M.F. | 81 | 39 (48.1%) | 7 (8.6%) | 10 (12.3%) | 27 (33.3%) | 30 (37.0%) | 37 (45.7%) | 67 (82.7%) | 74 (91.4%) |
| Symptom | 268 | 95 (35.4%) | 41 (15.3%) | 26 (9.7%) | 109 (40.7%) | 132 (49.3%) | 135 (50.4%) | 267 (99.6%) | 308 (114.9%) |
| Past H. | 518 | 343 (66.2%) | 45 (8.7%) | 18 (3.5%) | 113 (21.8%) | 122 (23.6%) | 131 (25.3%) | 253 (48.8%) | 298 (57.5%) |
| Fam. H. | 45 | 39 (86.7%) | 5 (11.1%) | 0 (0.0%) | 2 (4.4%) | 6 (13.3%) | 2 (4.4%) | *8 (17.8%)* | 13 (28.9%) |

| Social H. | 129 | 63 (48.8%) | 53 (41.1%) | 6 (4.7%) | 17 (13.2%) | 12 (9.3%) | 23 (17.8%) | 35 (27.1%) | 88 (68.2%) |
|---|---|---|---|---|---|---|---|---|---|
| *Total* | 1449 | 746 (51.5%) | 229 (15.8%) | 142 (9.8%) | 364 (25.1%) | 395 (27.3%) | 506 (34.9%) | 901 (62.2%) | 1130 (78.0%) |
| GatorTron + CLAMP | | | | | | | | | |
| Entity | Count | EM | RM | MM | UD | OD | MMUD | Error | RM+Error |
| CC | 133 | 72 (54.1%) | 12 (9.0%) | 31 (23.3%) | 19 (14.3%) | 20 (15.0%) | 50 (37.6%) | 70 (52.6%) | 82 (61.7%) |
| Location | 48 | 15 (31.3%) | 15 (31.3%) | 7 (14.6%) | 16 (33.3%) | 10 (20.8%) | 23 (47.9%) | 33 (68.8%) | 48 (100.0%) |
| Quality | 53 | 15 (28.3%) | 19 (35.8%) | 4 (7.5%) | 20 (37.7%) | 13 (24.5%) | 24 (45.3%) | ***37 (69.8%)*** | 56 (105.7%) |
| Severity | 34 | 4 (11.8%) | 8 (23.5%) | 4 (11.8%) | 20 (58.8%) | 10 (29.4%) | 24 (70.6%) | 34 (100.0%) | 42 (123.5%) |
| Duration | 66 | 37 (56.1%) | 18 (27.3%) | 1 (1.5%) | 12 (18.2%) | 21 (31.8%) | 13 (19.7%) | 34 (51.5%) | 52 (78.8%) |
| Timing | 37 | 14 (37.8%) | 3 (8.1%) | 9 (24.3%) | 12 (32.4%) | 11 (29.7%) | 21 (56.8%) | 32 (86.5%) | 35 (94.6%) |
| Context | 37 | 3 (8.1%) | 12 (32.4%) | 13 (35.1%) | 12 (32.4%) | 9 (24.3%) | 25 (67.6%) | 34 (91.9%) | 46 (124.3%) |
| M.F. | 81 | 36 (44.4%) | 10 (12.3%) | 8 (9.9%) | 29 (35.8%) | 26 (32.1%) | 37 (45.7%) | ***63 (77.8%)*** | 73 (90.1%) |
| Symptom | 268 | 77 (28.7%) | 37 (13.8%) | 35 (13.1%) | 124 (46.3%) | 94 (35.1%) | 159 (59.3%) | ***253 (94.4%)*** | 290 (108.2%) |
| Past H. | 518 | 347 (67.0%) | 39 (7.5%) | 11 (2.1%) | 123 (23.7%) | 118 (22.8%) | 134 (25.9%) | 252 (48.6%) | 291 (56.2%) |
| Fam. H. | 45 | 39 (86.7%) | 4 (8.9%) | 0 (0.0%) | 2 (4.4%) | 6 (13.3%) | 2 (4.4%) | ***8 (17.8%)*** | 12 (26.7%) |
| Social H. | 129 | 61 (47.3%) | 55 (42.6%) | 6 (4.7%) | 20 (15.5%) | 13 (10.1%) | 26 (20.2%) | 39 (30.2%) | 94 (72.9%) |
| *Total* | 1449 | 720 (49.7%) | 232 (16.0%) | 129 (8.9%) | 409 (28.2%) | 351 (24.2%) | 538 (37.1%) | 889 (61.4%) | 1121 (77.4%) |
| GatorTronS | | | | | | | | | |
| Entity | Count | EM | RM | MM | UD | OD | MMUD | Error | RM+Error |
| CC | 133 | 67 (50.4%) | 11 (8.3%) | 25 (18.8%) | 30 (22.6%) | 24 (18.0%) | 55 (41.4%) | 79 (59.4%) | 90 (67.7%) |
| Location | 48 | 14 (29.2%) | 13 (27.1%) | 8 (16.7%) | 17 (35.4%) | 10 (20.8%) | 25 (52.1%) | 35 (72.9%) | 48 (100.0%) |
| Quality | 53 | 19 (35.8%) | 11 (20.8%) | 4 (7.5%) | 21 (39.6%) | 14 (26.4%) | 25 (47.2%) | 39 (73.6%) | 50 (94.3%) |
| Severity | 34 | 5 (14.7%) | 6 (17.6%) | 8 (23.5%) | 16 (47.1%) | 6 (17.6%) | 24 (70.6%) | ***30 (88.2%)*** | 36 (105.9%) |
| Duration | 66 | 34 (51.5%) | 19 (28.8%) | 0 (0.0%) | 18 (27.3%) | 21 (31.8%) | 18 (27.3%) | 39 (59.1%) | 58 (87.9%) |
| Timing | 37 | 13 (35.1%) | 1 (2.7%) | 9 (24.3%) | 14 (37.8%) | 10 (27.0%) | 23 (62.2%) | 33 (89.2%) | 34 (91.9%) |
| Context | 37 | 2 (5.4%) | 11 (29.7%) | 13 (35.1%) | 13 (35.1%) | 12 (32.4%) | 26 (70.3%) | 38 (102.7%) | 49 (132.4%) |
| M.F. | 81 | 32 (39.5%) | 4 (4.9%) | 7 (8.6%) | 39 (48.1%) | 32 (39.5%) | 46 (56.8%) | 78 (96.3%) | 82 (101.2%) |
| Symptom | 268 | 92 (34.3%) | 33 (12.3%) | 27 (10.1%) | 120 (44.8%) | 125 (46.6%) | 147 (54.9%) | 272 (101.5%) | 305 (113.8%) |
| Past H. | 518 | 341 (65.8%) | 40 (7.7%) | 14 (2.7%) | 124 (23.9%) | 126 (24.3%) | 138 (26.6%) | 264 (51.0%) | 304 (58.7%) |
| Fam. H. | 45 | 36 (80.0%) | 5 (11.1%) | 1 (2.2%) | 4 (8.9%) | 5 (11.1%) | 5 (11.1%) | 10 (22.2%) | 15 (33.3%) |
| Social H. | 129 | 64 (49.6%) | 55 (42.6%) | 4 (3.1%) | 17 (13.2%) | 16 (12.4%) | 21 (16.3%) | 37 (28.7%) | 92 (71.3%) |

| Total | 1449 | 719 (49.6%) | 209 (14.4%) | 120 (8.3%) | 433 (29.9%) | 401 (27.7%) | 553 (38.2%) | 954 (65.8%) | 1163 (80.3%) |
|---|---|---|---|---|---|---|---|---|---|
| | | | | GatorTronS + CLAMP | | | | | |
| Entity | Count | EM | RM | MM | UD | OD | MMUD | Error | RM+Error |
| CC | 133 | 75 (56.4%) | 14 (10.5%) | 24 (18.0%) | 22 (16.5%) | 19 (14.3%) | 46 (34.6%) | ***65 (48.9%)*** | 79 (59.4%) |
| Location | 48 | 13 (27.1%) | 10 (20.8%) | 8 (16.7%) | 19 (39.6%) | 5 (10.4%) | 27 (56.3%) | ***32 (66.7%)*** | 42 (87.5%) |
| Quality | 53 | 15 (28.3%) | 8 (15.1%) | 7 (13.2%) | 25 (47.2%) | 9 (17.0%) | 32 (60.4%) | 41 (77.4%) | 49 (92.5%) |
| Severity | 34 | 1 (2.9%) | 2 (5.9%) | 6 (17.6%) | 25 (73.5%) | 2 (5.9%) | 31 (91.2%) | 33 (97.1%) | 35 (102.9%) |
| Duration | 66 | 35 (53.0%) | 15 (22.7%) | 0 (0.0%) | 19 (28.8%) | 14 (21.2%) | 19 (28.8%) | ***33 (50.0%)*** | 48 (72.7%) |
| Timing | 37 | 12 (32.4%) | 2 (5.4%) | 8 (21.6%) | 15 (40.5%) | 9 (24.3%) | 23 (62.2%) | 32 (86.5%) | 34 (91.9%) |
| Context | 37 | 1 (2.7%) | 7 (18.9%) | 15 (40.5%) | 15 (40.5%) | 8 (21.6%) | 30 (81.1%) | 38 (102.7%) | 45 (121.6%) |
| M.F. | 81 | 31 (38.3%) | 6 (7.4%) | 9 (11.1%) | 35 (43.2%) | 28 (34.6%) | 44 (54.3%) | 72 (88.9%) | 78 (96.3%) |
| Symptom | 268 | 78 (29.1%) | 40 (14.9%) | 23 (8.6%) | 132 (49.3%) | 110 (41.0%) | 155 (57.8%) | 265 (98.9%) | 305 (113.8%) |
| Past H. | 518 | 343 (66.2%) | 48 (9.3%) | 12 (2.3%) | 117 (22.6%) | 101 (19.5%) | 129 (24.9%) | ***230 (44.4%)*** | 278 (53.7%) |
| Fam. H. | 45 | 38 (84.4%) | 3 (6.7%) | 0 (0.0%) | 4 (8.9%) | 5 (11.1%) | 4 (8.9%) | 9 (20.0%) | 12 (26.7%) |
| Social H. | 129 | 56 (43.4%) | 49 (38.0%) | 2 (1.6%) | 31 (24.0%) | 13 (10.1%) | 33 (25.6%) | 46 (35.7%) | 95 (73.6%) |
| Total | 1449 | 698 (48.2%) | 204 (14.1%) | 114 (7.9%) | 459 (31.7%) | 323 (22.3%) | 573 (39.5%) | 896 (61.8%) | 1100 (75.9%) |
| | | | | BioMegatron | | | | | |
| Entity | Count | EM | RM | MM | UD | OD | MMUD | Error | RM+Error |
| CC | 133 | 57 (42.9%) | 8 (6.0%) | 36 (27.1%) | 33 (24.8%) | 22 (16.5%) | 69 (51.9%) | 91 (68.4%) | 99 (74.4%) |
| Location | 48 | 12 (25.0%) | 12 (25.0%) | 8 (16.7%) | 19 (39.6%) | 21 (43.8%) | 27 (56.3%) | 48 (100.0%) | 60 (125.0%) |
| Quality | 53 | 10 (18.9%) | 9 (17.0%) | 7 (13.2%) | 29 (54.7%) | 10 (18.9%) | 36 (67.9%) | 46 (86.8%) | 55 (103.8%) |
| Severity | 34 | 3 (8.8%) | 6 (17.6%) | 6 (17.6%) | 20 (58.8%) | 8 (23.5%) | 26 (76.5%) | 34 (100.0%) | 40 (117.6%) |
| Duration | 66 | 30 (45.5%) | 19 (28.8%) | 1 (1.5%) | 20 (30.3%) | 29 (43.9%) | 21 (31.8%) | 50 (75.8%) | 69 (104.5%) |
| Timing | 37 | 14 (37.8%) | 1 (2.7%) | 11 (29.7%) | 11 (29.7%) | 12 (32.4%) | 22 (59.5%) | 34 (91.9%) | 35 (94.6%) |
| Context | 37 | 0 (0.0%) | 7 (18.9%) | 20 (54.1%) | 12 (32.4%) | 10 (27.0%) | 32 (86.5%) | 42 (113.5%) | 49 (132.4%) |
| M.F. | 81 | 30 (37.0%) | 8 (9.9%) | 13 (16.0%) | 33 (40.7%) | 45 (55.6%) | 46 (56.8%) | 91 (112.3%) | 99 (122.2%) |
| Symptom | 268 | 80 (29.9%) | 37 (13.8%) | 39 (14.6%) | 118 (44.0%) | 140 (52.2%) | 157 (58.6%) | 297 (110.8%) | 334 (124.6%) |
| Past H. | 518 | 344 (66.4%) | 48 (9.3%) | 15 (2.9%) | 117 (22.6%) | 167 (32.2%) | 132 (25.5%) | 299 (57.7%) | 347 (67.0%) |
| Fam. H. | 45 | 35 (77.8%) | 4 (8.9%) | 2 (4.4%) | 5 (11.1%) | 3 (6.7%) | 7 (15.6%) | 10 (22.2%) | 14 (31.1%) |
| Social H. | 129 | 55 (42.6%) | 50 (38.8%) | 7 (5.4%) | 24 (18.6%) | 10 (7.8%) | 31 (24.0%) | 41 (31.8%) | 91 (70.5%) |
| Total | 1449 | 670 (46.2%) | 209 (14.4%) | 165 (11.4%) | 441 (30.4%) | 477 (32.9%) | 606 (41.8%) | 1083 (74.7%) | 1292 (89.2%) |

| BioMegatron + CLAMP | | | | | | | | | |
|---|---|---|---|---|---|---|---|---|---|
| Entity | Count | EM | RM | MM | UD | OD | MMUD | Error | RM+Error |
| CC | 133 | 64 (48.1%) | 15 (11.3%) | 32 (24.1%) | 25 (18.8%) | 26 (19.5%) | 57 (42.9%) | 83 (62.4%) | 98 (73.7%) |
| Location | 48 | 10 (20.8%) | 11 (22.9%) | 7 (14.6%) | 22 (45.8%) | 16 (33.3%) | 29 (60.4%) | 45 (93.8%) | 56 (116.7%) |
| Quality | 53 | 10 (18.9%) | 10 (18.9%) | 8 (15.1%) | 28 (52.8%) | 11 (20.8%) | 36 (67.9%) | 47 (88.7%) | 57 (107.5%) |
| Severity | 34 | 2 (5.9%) | 3 (8.8%) | 7 (20.6%) | 22 (64.7%) | 1 (2.9%) | 29 (85.3%) | 30 (88.2%) | 33 (97.1%) |
| Duration | 66 | 32 (48.5%) | 16 (24.2%) | 1 (1.5%) | 19 (28.8%) | 22 (33.3%) | 20 (30.3%) | 42 (63.6%) | 58 (87.9%) |
| Timing | 37 | 11 (29.7%) | 0 (0.0%) | 8 (21.6%) | 18 (48.6%) | 13 (35.1%) | 26 (70.3%) | 39 (105.4%) | 39 (105.4%) |
| Context | 37 | 0 (0.0%) | 2 (5.4%) | 17 (45.9%) | 18 (48.6%) | 4 (10.8%) | 35 (94.6%) | 39 (105.4%) | 41 (110.8%) |
| M.F. | 81 | 25 (30.9%) | 10 (12.3%) | 9 (11.1%) | 38 (46.9%) | 42 (51.9%) | 47 (58.0%) | 89 (109.9%) | 99 (122.2%) |
| Symptom | 268 | 57 (21.3%) | 33 (12.3%) | 34 (12.7%) | 145 (54.1%) | 127 (47.4%) | 179 (66.8%) | 306 (114.2%) | 339 (126.5%) |
| Past H. | 518 | 342 (66.0%) | 46 (8.9%) | 9 (1.7%) | 127 (24.5%) | 157 (30.3%) | 136 (26.3%) | 293 (56.6%) | 339 (65.4%) |
| Fam. H. | 45 | 32 (71.1%) | 3 (6.7%) | 5 (11.1%) | 5 (11.1%) | 6 (13.3%) | 10 (22.2%) | 16 (35.6%) | 19 (42.2%) |
| Social H. | 129 | 57 (44.2%) | 53 (41.1%) | 5 (3.9%) | 25 (19.4%) | 20 (15.5%) | 30 (23.3%) | 50 (38.8%) | 103 (79.8%) |
| *Total* | 1449 | 642 (44.3%) | 202 (13.9%) | 142 (9.8%) | 492 (34.0%) | 445 (30.7%) | 634 (43.8%) | 1079 (74.5%) | 1281 (88.4%) |
| PubMedBERT | | | | | | | | | |
| Entity | Count | EM | RM | MM | UD | OD | MMUD | Error | RM+Error |
| CC | 133 | 47 (35.3%) | 11 (8.3%) | 47 (35.3%) | 29 (21.8%) | 8 (6.0%) | 76 (57.1%) | 84 (63.2%) | 95 (71.4%) |
| Location | 48 | 10 (20.8%) | 7 (14.6%) | 11 (22.9%) | 21 (43.8%) | 15 (31.3%) | 32 (66.7%) | 47 (97.9%) | 54 (112.5%) |
| Quality | 53 | 14 (26.4%) | 8 (15.1%) | 8 (15.1%) | 23 (43.4%) | 18 (34.0%) | 31 (58.5%) | 49 (92.5%) | 57 (107.5%) |
| Severity | 34 | 4 (11.8%) | 3 (8.8%) | 14 (41.2%) | 14 (41.2%) | 5 (14.7%) | 28 (82.4%) | 33 (97.1%) | 36 (105.9%) |
| Duration | 66 | 28 (42.4%) | 9 (13.6%) | 1 (1.5%) | 28 (42.4%) | 33 (50.0%) | 29 (43.9%) | 62 (93.9%) | 71 (107.6%) |
| Timing | 37 | 12 (32.4%) | 0 (0.0%) | 12 (32.4%) | 13 (35.1%) | 16 (43.2%) | 25 (67.6%) | 41 (110.8%) | 41 (110.8%) |
| Context | 37 | 0 (0.0%) | 9 (24.3%) | 20 (54.1%) | 10 (27.0%) | 9 (24.3%) | 30 (81.1%) | 39 (105.4%) | 48 (129.7%) |
| M.F. | 81 | 34 (42.0%) | 3 (3.7%) | 13 (16.0%) | 32 (39.5%) | 36 (44.4%) | 45 (55.6%) | 81 (100.0%) | 84 (103.7%) |
| Symptom | 268 | 80 (29.9%) | 38 (14.2%) | 30 (11.2%) | 123 (45.9%) | 158 (59.0%) | 153 (57.1%) | 311 (116.0%) | 349 (130.2%) |
| Past H. | 518 | 355 (68.5%) | 40 (7.7%) | 18 (3.5%) | 108 (20.8%) | 173 (33.4%) | 126 (24.3%) | 299 (57.7%) | 339 (65.4%) |
| Fam. H. | 45 | 35 (77.8%) | 3 (6.7%) | 4 (8.9%) | 3 (6.7%) | 3 (6.7%) | 7 (15.6%) | 10 (22.2%) | 13 (28.9%) |
| Social H. | 129 | 60 (46.5%) | 49 (38.0%) | 5 (3.9%) | 23 (17.8%) | 26 (20.2%) | 28 (21.7%) | 54 (41.9%) | 103 (79.8%) |
| *Total* | 1449 | 679 (46.9%) | 180 (12.4%) | 183 (12.6%) | 427 (29.5%) | 500 (34.5%) | 610 (42.1%) | 1110 (76.6%) | 1290 (89.0%) |
| PubMedBERT + CLAMP | | | | | | | | | |

| Entity | Count | EM | RM | MM | UD | OD | MMUD | Error | RM+Error |
|---|---|---|---|---|---|---|---|---|---|
| CC | 133 | 62 (46.6%) | 16 (12.0%) | 35 (26.3%) | 23 (17.3%) | 25 (18.8%) | 58 (43.6%) | 83 (62.4%) | 99 (74.4%) |
| Location | 48 | 9 (18.8%) | 7 (14.6%) | 11 (22.9%) | 22 (45.8%) | 21 (43.8%) | 33 (68.8%) | 54 (112.5%) | 61 (127.1%) |
| Quality | 53 | 10 (18.9%) | 10 (18.9%) | 10 (18.9%) | 26 (49.1%) | 15 (28.3%) | 36 (67.9%) | 51 (96.2%) | 61 (115.1%) |
| Severity | 34 | 1 (2.9%) | 5 (14.7%) | 10 (29.4%) | 19 (55.9%) | 6 (17.6%) | 29 (85.3%) | 35 (102.9%) | 40 (117.6%) |
| Duration | 66 | 27 (40.9%) | 11 (16.7%) | 2 (3.0%) | 27 (40.9%) | 34 (51.5%) | 29 (43.9%) | 63 (95.5%) | 74 (112.1%) |
| Timing | 37 | 7 (18.9%) | 0 (0.0%) | 14 (37.8%) | 16 (43.2%) | 9 (24.3%) | 30 (81.1%) | 39 (105.4%) | 39 (105.4%) |
| Context | 37 | 1 (2.7%) | 4 (10.8%) | 18 (48.6%) | 15 (40.5%) | 4 (10.8%) | 33 (89.2%) | 37 (100.0%) | 41 (110.8%) |
| M.F. | 81 | 23 (28.4%) | 7 (8.6%) | 15 (18.5%) | 39 (48.1%) | 31 (38.3%) | 54 (66.7%) | 85 (104.9%) | 92 (113.6%) |
| Symptom | 268 | 63 (23.5%) | 27 (10.1%) | 39 (14.6%) | 141 (52.6%) | 128 (47.8%) | 180 (67.2%) | 308 (114.9%) | 335 (125.0%) |
| Past H. | 518 | 348 (67.2%) | 44 (8.5%) | 15 (2.9%) | 115 (22.2%) | 171 (33.0%) | 130 (25.1%) | 301 (58.1%) | 345 (66.6%) |
| Fam. H. | 45 | 33 (73.3%) | 1 (2.2%) | 8 (17.8%) | 3 (6.7%) | 5 (11.1%) | 11 (24.4%) | 16 (35.6%) | 17 (37.8%) |
| Social H. | 129 | 55 (42.6%) | 62 (48.1%) | 5 (3.9%) | 23 (17.8%) | 25 (19.4%) | 28 (21.7%) | 53 (41.1%) | 115 (89.1%) |
| *Total* | 1449 | 639 (44.1%) | 194 (13.4%) | 182 (12.6%) | 469 (32.4%) | 474 (32.7%) | 651 (44.9%) | 1125 (77.6%) | 1319 (91.0%) |
| PubMedBERT Large | | | | | | | | | |
| Entity | Count | EM | RM | MM | UD | OD | MMUD | Error | RM+Error |
| CC | 133 | 56 (42.1%) | 16 (12.0%) | 37 (27.8%) | 27 (20.3%) | 17 (12.8%) | 64 (48.1%) | 81 (60.9%) | 97 (72.9%) |
| Location | 48 | 13 (27.1%) | 11 (22.9%) | 10 (20.8%) | 17 (35.4%) | 18 (37.5%) | 27 (56.3%) | 45 (93.8%) | 56 (116.7%) |
| Quality | 53 | 11 (20.8%) | 6 (11.3%) | 10 (18.9%) | 27 (50.9%) | 17 (32.1%) | 37 (69.8%) | 54 (101.9%) | 60 (113.2%) |
| Severity | 34 | 2 (5.9%) | 1 (2.9%) | 17 (50.0%) | 14 (41.2%) | 4 (11.8%) | 31 (91.2%) | 35 (102.9%) | 36 (105.9%) |
| Duration | 66 | 31 (47.0%) | 10 (15.2%) | 2 (3.0%) | 24 (36.4%) | 32 (48.5%) | 26 (39.4%) | 58 (87.9%) | 68 (103.0%) |
| Timing | 37 | 10 (27.0%) | 1 (2.7%) | 14 (37.8%) | 12 (32.4%) | 10 (27.0%) | 26 (70.3%) | 36 (97.3%) | 37 (100.0%) |
| Context | 37 | 0 (0.0%) | 8 (21.6%) | 18 (48.6%) | 12 (32.4%) | 11 (29.7%) | 30 (81.1%) | 41 (110.8%) | 49 (132.4%) |
| M.F. | 81 | 31 (38.3%) | 5 (6.2%) | 9 (11.1%) | 37 (45.7%) | 53 (65.4%) | 46 (56.8%) | 99 (122.2%) | 104 (128.4%) |
| Symptom | 268 | 83 (31.0%) | 36 (13.4%) | 30 (11.2%) | 123 (45.9%) | 125 (46.6%) | 153 (57.1%) | 278 (103.7%) | 314 (117.2%) |
| Past H. | 518 | 340 (65.6%) | 45 (8.7%) | 17 (3.3%) | 119 (23.0%) | 161 (31.1%) | 136 (26.3%) | 297 (57.3%) | 342 (66.0%) |
| Fam. H. | 45 | 36 (80.0%) | 5 (11.1%) | 3 (6.7%) | 3 (6.7%) | 4 (8.9%) | 6 (13.3%) | 10 (22.2%) | 15 (33.3%) |
| Social H. | 129 | 58 (45.0%) | 56 (43.4%) | 6 (4.7%) | 21 (16.3%) | 7 (5.4%) | 27 (20.9%) | *34 (26.4%)* | 90 (69.8%) |
| *Total* | 1449 | 671 (46.3%) | 200 (13.8%) | 173 (11.9%) | 436 (30.1%) | 459 (31.7%) | 609 (42.0%) | 1068 (73.7%) | 1268 (87.5%) |
| PubMedBERT Large + CLAMP | | | | | | | | | |
| Entity | Count | EM | RM | MM | UD | OD | MMUD | Error | RM+Error |
| CC | 133 | 62 (46.6%) | 13 (9.8%) | 33 (24.8%) | 27 (20.3%) | 16 (12.0%) | 60 (45.1%) | 76 (57.1%) | 89 (66.9%) |
| Location | 48 | 13 (27.1%) | 5 (10.4%) | 7 (14.6%) | 25 (52.1%) | 8 (16.7%) | 32 (66.7%) | 40 (83.3%) | 45 (93.8%) |
| Quality | 53 | 9 (17.0%) | 9 (17.0%) | 10 (18.9%) | 27 (50.9%) | 13 (24.5%) | 37 (69.8%) | 50 (94.3%) | 59 (111.3%) |
| Severity | 34 | 0 (0.0%) | 2 (5.9%) | 10 (29.4%) | 22 (64.7%) | 6 (17.6%) | 32 (94.1%) | 38 (111.8%) | 40 (117.6%) |
| Duration | 66 | 27 (40.9%) | 15 (22.7%) | 2 (3.0%) | 22 (33.3%) | 28 (42.4%) | 24 (36.4%) | 52 (78.8%) | 67 (101.5%) |
| Timing | 37 | 16 (43.2%) | 0 (0.0%) | 10 (27.0%) | 11 (29.7%) | 10 (27.0%) | 21 (56.8%) | 31 (83.8%) | 31 (83.8%) |
| Context | 37 | 0 (0.0%) | 11 (29.7%) | 20 (54.1%) | 7 (18.9%) | 12 (32.4%) | 27 (73.0%) | 39 (105.4%) | 50 (135.1%) |
| M.F. | 81 | 33 (40.7%) | 2 (2.5%) | 3 (3.7%) | 43 (53.1%) | 37 (45.7%) | 46 (56.8%) | 83 (102.5%) | 85 (104.9%) |
| Symptom | 268 | 76 (28.4%) | 30 (11.2%) | 27 (10.1%) | 141 (52.6%) | 133 (49.6%) | 168 (62.7%) | 301 (112.3%) | 331 (123.5%) |
| Past H. | 518 | 336 (64.9%) | 40 (7.7%) | 19 (3.7%) | 126 (24.3%) | 150 (29.0%) | 145 (28.0%) | 295 (56.9%) | 335 (64.7%) |
| Fam. H. | 45 | 36 (80.0%) | 3 (6.7%) | 3 (6.7%) | 4 (8.9%) | 4 (8.9%) | 7 (15.6%) | 11 (24.4%) | 14 (31.1%) |
| Social H. | 129 | 57 (44.2%) | 53 (41.1%) | 6 (4.7%) | 25 (19.4%) | 13 (10.1%) | 31 (24.0%) | 44 (34.1%) | 97 (75.2%) |
| *Total* | 1449 | 665 (45.9%) | 183 (12.6%) | 150 (10.4%) | 480 (33.1%) | 430 (29.7%) | 630 (43.5%) | 1060 (73.2%) | 1243 (85.8%) |

Notes: Abbreviations: EM (Exact Match), RM (Relaxed Match), MM(Mismatch), UD (Under Detection), OD (Over Detection). Highlights: Integration of BMEs identified by CLAMP reduced the errors. ***Red italic bold font***: The lowest error rate among all models assessed for the concept.

Table S3: Detailed counts and rates of matching results for each sample for GatorTron/GatorTronS-based models.

| | | | GatorTronS | | | | | | | |
|---|---|---|---|---|---|---|---|---|---|---|
| Sample | Anno # | Word # | EM | RM | MM | UD | OD | MMUD | Error | RM+Error |
| sample_1128 | 33 | 492 | 18 (54.5%) | 2 (6.1%) | 2 (6.1%) | 11 (33.3%) | 0 (0.0%) | 13 (39.4%) | 13 (39.4%) | 15 (45.5%) |
| sample_1133 | 24 | 782 | 8 (33.3%) | 2 (8.3%) | 1 (4.2%) | 13 (54.2%) | 1 (4.2%) | 14 (58.3%) | 15 (62.5%) | 17 (70.8%) |
| sample_1152 | 18 | 593 | 10 (55.6%) | 0 (0.0%) | 4 (22.2%) | 4 (22.2%) | 2 (11.1%) | 8 (44.4%) | 10 (55.6%) | 10 (55.6%) |
| sample_1169 | 8 | 422 | 4 (50.0%) | 0 (0.0%) | 1 (12.5%) | 3 (37.5%) | 1 (12.5%) | 4 (50.0%) | 5 (62.5%) | 5 (62.5%) |
| sample_1242 | 11 | 536 | 3 (27.3%) | 0 (0.0%) | 1 (9.1%) | 7 (63.6%) | 4 (36.4%) | 8 (72.7%) | 12 (109.1%) | 12 (109.1%) |
| sample_1248 | 31 | 910 | 13 (41.9%) | 1 (3.2%) | 5 (16.1%) | 12 (38.7%) | 16 (51.6%) | 17 (54.8%) | 33 (106.5%) | 34 (109.7%) |
| sample_1252 | 19 | 585 | 11 (57.9%) | 5 (26.3%) | 2 (10.5%) | 2 (10.5%) | 3 (15.8%) | 4 (21.1%) | 7 (36.8%) | 12 (63.2%) |
| sample_1419 | 10 | 540 | 3 (30.0%) | 4 (40.0%) | 2 (20.0%) | 2 (20.0%) | 1 (10.0%) | 4 (40.0%) | 5 (50.0%) | 9 (90.0%) |
| sample_1439 | 19 | 896 | 3 (15.8%) | 2 (10.5%) | 4 (21.1%) | 10 (52.6%) | 1 (5.3%) | 14 (73.7%) | 15 (78.9%) | 17 (89.5%) |
| sample_1495 | 45 | 872 | 17 (37.8%) | 0 (0.0%) | 2 (4.4%) | 26 (57.8%) | 1 (2.2%) | 28 (62.2%) | 29 (64.4%) | 29 (64.4%) |
| sample_1505 | 16 | 784 | 6 (37.5%) | 3 (18.8%) | 3 (18.8%) | 4 (25.0%) | 6 (37.5%) | 7 (43.8%) | 13 (81.3%) | 16 (100.0%) |

| | | | | | | | | | |
|---|---|---|---|---|---|---|---|---|---|
| sample_1568 | 28 | 1112 | 19 (67.9%) | 3 (10.7%) | 2 (7.1%) | 4 (14.3%) | 13 (46.4%) | 6 (21.4%) | 19 (67.9%) | 22 (78.6%) |
| sample_1592 | 24 | 785 | 16 (66.7%) | 2 (8.3%) | 0 (0.0%) | 6 (25.0%) | 4 (16.7%) | 6 (25.0%) | 10 (41.7%) | 12 (50.0%) |
| sample_1921 | 21 | 789 | 14 (66.7%) | 3 (14.3%) | 1 (4.8%) | 3 (14.3%) | 10 (47.6%) | 4 (19.0%) | 14 (66.7%) | 17 (81.0%) |
| sample_1956 | 14 | 628 | 9 (64.3%) | 1 (7.1%) | 2 (14.3%) | 2 (14.3%) | 13 (92.9%) | 4 (28.6%) | 17 (121.4%) | 18 (128.6%) |
| sample_2129 | 13 | 427 | 4 (30.8%) | 4 (30.8%) | 1 (7.7%) | 4 (30.8%) | 12 (92.3%) | 5 (38.5%) | 17 (130.8%) | 21 (161.5%) |
| sample_214 | 14 | 586 | 5 (35.7%) | 4 (28.6%) | 4 (28.6%) | 2 (14.3%) | 6 (42.9%) | 6 (42.9%) | 12 (85.7%) | 16 (114.3%) |
| sample_2210 | 38 | 874 | 6 (15.8%) | 7 (18.4%) | 0 (0.0%) | 27 (71.1%) | 7 (18.4%) | 27 (71.1%) | 34 (89.5%) | 41 (107.9%) |
| sample_2218 | 35 | 771 | 18 (51.4%) | 12 (34.3%) | 2 (5.7%) | 5 (14.3%) | 10 (28.6%) | 7 (20.0%) | 17 (48.6%) | 29 (82.9%) |
| sample_223 | 11 | 567 | 4 (36.4%) | 2 (18.2%) | 2 (18.2%) | 4 (36.4%) | 7 (63.6%) | 6 (54.5%) | 13 (118.2%) | 15 (136.4%) |
| sample_225 | 13 | 468 | 4 (30.8%) | 4 (30.8%) | 1 (7.7%) | 4 (30.8%) | 3 (23.1%) | 5 (38.5%) | 8 (61.5%) | 12 (92.3%) |
| sample_226 | 8 | 1022 | 3 (37.5%) | 2 (25.0%) | 1 (12.5%) | 2 (25.0%) | 19 (237.5%) | 3 (37.5%) | 22 (275.0%) | 24 (300.0%) |
| sample_2275 | 27 | 612 | 8 (29.6%) | 2 (7.4%) | 5 (18.5%) | 12 (44.4%) | 3 (11.1%) | 17 (63.0%) | 20 (74.1%) | 22 (81.5%) |
| sample_2604 | 20 | 733 | 5 (25.0%) | 5 (25.0%) | 3 (15.0%) | 8 (40.0%) | 5 (25.0%) | 11 (55.0%) | 16 (80.0%) | 21 (105.0%) |
| sample_2623 | 14 | 480 | 6 (42.9%) | 1 (7.1%) | 2 (14.3%) | 5 (35.7%) | 10 (71.4%) | 7 (50.0%) | 17 (121.4%) | 18 (128.6%) |
| sample_2746 | 31 | 2066 | 24 (77.4%) | 3 (9.7%) | 1 (3.2%) | 3 (9.7%) | 18 (58.1%) | 4 (12.9%) | 22 (71.0%) | 25 (80.6%) |
| sample_2747 | 13 | 814 | 5 (38.5%) | 2 (15.4%) | 0 (0.0%) | 6 (46.2%) | 3 (23.1%) | 6 (46.2%) | 9 (69.2%) | 11 (84.6%) |
| sample_2780 | 48 | 994 | 20 (41.7%) | 17 (35.4%) | 6 (12.5%) | 13 (27.1%) | 6 (12.5%) | 19 (39.6%) | 25 (52.1%) | 42 (87.5%) |
| sample_2789 | 34 | 1082 | 26 (76.5%) | 5 (14.7%) | 1 (2.9%) | 3 (8.8%) | 18 (52.9%) | 4 (11.8%) | 22 (64.7%) | 27 (79.4%) |
| sample_2790 | 34 | 1123 | 21 (61.8%) | 5 (14.7%) | 3 (8.8%) | 6 (17.6%) | 24 (70.6%) | 9 (26.5%) | 33 (97.1%) | 38 (111.8%) |
| sample_2792 | 9 | 708 | 5 (55.6%) | 1 (11.1%) | 1 (11.1%) | 2 (22.2%) | 27 (300.0%) | 3 (33.3%) | 30 (333.3%) | 31 (344.4%) |
| sample_343 | 25 | 1075 | 13 (52.0%) | 3 (12.0%) | 4 (16.0%) | 5 (20.0%) | 11 (44.0%) | 9 (36.0%) | 20 (80.0%) | 23 (92.0%) |
| sample_365 | 57 | 1312 | 41 (71.9%) | 8 (14.0%) | 5 (8.8%) | 4 (7.0%) | 12 (21.1%) | 9 (15.8%) | 21 (36.8%) | 29 (50.9%) |
| sample_377 | 55 | 1420 | 22 (40.0%) | 9 (16.4%) | 9 (16.4%) | 16 (29.1%) | 11 (20.0%) | 25 (45.5%) | 36 (65.5%) | 45 (81.8%) |
| sample_378 | 24 | 939 | 10 (41.7%) | 2 (8.3%) | 2 (8.3%) | 10 (41.7%) | 8 (33.3%) | 12 (50.0%) | 20 (83.3%) | 22 (91.7%) |
| sample_380 | 50 | 1139 | 22 (44.0%) | 2 (4.0%) | 2 (4.0%) | 24 (48.0%) | 5 (10.0%) | 26 (52.0%) | 31 (62.0%) | 33 (66.0%) |
| sample_388 | 17 | 620 | 6 (35.3%) | 4 (23.5%) | 2 (11.8%) | 5 (29.4%) | 10 (58.8%) | 7 (41.2%) | 17 (100.0%) | 21 (123.5%) |
| sample_391 | 20 | 918 | 6 (30.0%) | 1 (5.0%) | 3 (15.0%) | 10 (50.0%) | 8 (40.0%) | 13 (65.0%) | 21 (105.0%) | 22 (110.0%) |
| sample_392 | 34 | 1004 | 14 (41.2%) | 2 (5.9%) | 2 (5.9%) | 16 (47.1%) | 3 (8.8%) | 18 (52.9%) | 21 (61.8%) | 23 (67.6%) |
| sample_393 | 50 | 1552 | 28 (56.0%) | 6 (12.0%) | 0 (0.0%) | 16 (32.0%) | 6 (12.0%) | 16 (32.0%) | 22 (44.0%) | 28 (56.0%) |
| sample_394 | 19 | 676 | 8 (42.1%) | 0 (0.0%) | 5 (26.3%) | 6 (31.6%) | 1 (5.3%) | 11 (57.9%) | 12 (63.2%) | 12 (63.2%) |
| sample_398 | 24 | 694 | 17 (70.8%) | 3 (12.5%) | 0 (0.0%) | 4 (16.7%) | 4 (16.7%) | 4 (16.7%) | 8 (33.3%) | 11 (45.8%) |

| Sample | Anno # | Word # | EM | RM | MM | UD | OD | MMUD | Error | RM+Error |
|---|---|---|---|---|---|---|---|---|---|---|
| sample_402 | 15 | 491 | 9 (60.0%) | 1 (6.7%) | 3 (20.0%) | 2 (13.3%) | 2 (13.3%) | 5 (33.3%) | 7 (46.7%) | 8 (53.3%) |
| sample_403 | 17 | 562 | 10 (58.8%) | 6 (35.3%) | 0 (0.0%) | 2 (11.8%) | 3 (17.6%) | 2 (11.8%) | 5 (29.4%) | 11 (64.7%) |
| sample_439 | 16 | 411 | 5 (31.3%) | 2 (12.5%) | 1 (6.3%) | 8 (50.0%) | 0 (0.0%) | 9 (56.3%) | 9 (56.3%) | 11 (68.8%) |
| sample_452 | 15 | 398 | 10 (66.7%) | 2 (13.3%) | 1 (6.7%) | 2 (13.3%) | 2 (13.3%) | 3 (20.0%) | 5 (33.3%) | 7 (46.7%) |
| sample_476 | 17 | 666 | 7 (41.2%) | 3 (17.6%) | 2 (11.8%) | 5 (29.4%) | 5 (29.4%) | 7 (41.2%) | 12 (70.6%) | 15 (88.2%) |
| sample_485 | 24 | 742 | 10 (41.7%) | 3 (12.5%) | 2 (8.3%) | 10 (41.7%) | 3 (12.5%) | 12 (50.0%) | 15 (62.5%) | 18 (75.0%) |
| sample_570 | 31 | 1038 | 16 (51.6%) | 9 (29.0%) | 3 (9.7%) | 5 (16.1%) | 12 (38.7%) | 8 (25.8%) | 20 (64.5%) | 29 (93.5%) |
| sample_579 | 30 | 932 | 19 (63.3%) | 2 (6.7%) | 3 (10.0%) | 6 (20.0%) | 7 (23.3%) | 9 (30.0%) | 16 (53.3%) | 18 (60.0%) |
| sample_583 | 29 | 927 | 15 (51.7%) | 6 (20.7%) | 0 (0.0%) | 8 (27.6%) | 9 (31.0%) | 8 (27.6%) | 17 (58.6%) | 23 (79.3%) |
| sample_664 | 19 | 885 | 13 (68.4%) | 1 (5.3%) | 0 (0.0%) | 5 (26.3%) | 2 (10.5%) | 5 (26.3%) | 7 (36.8%) | 8 (42.1%) |
| sample_666 | 40 | 873 | 22 (55.0%) | 6 (15.0%) | 3 (7.5%) | 11 (27.5%) | 4 (10.0%) | 14 (35.0%) | 18 (45.0%) | 24 (60.0%) |
| sample_687 | 20 | 426 | 7 (35.0%) | 8 (40.0%) | 1 (5.0%) | 7 (35.0%) | 0 (0.0%) | 8 (40.0%) | 8 (40.0%) | 16 (80.0%) |
| sample_70 | 7 | 519 | 3 (42.9%) | 0 (0.0%) | 0 (0.0%) | 4 (57.1%) | 0 (0.0%) | 4 (57.1%) | 4 (57.1%) | 4 (57.1%) |
| sample_71 | 12 | 562 | 3 (25.0%) | 3 (25.0%) | 0 (0.0%) | 7 (58.3%) | 2 (16.7%) | 7 (58.3%) | 9 (75.0%) | 12 (100.0%) |
| sample_782 | 10 | 355 | 2 (20.0%) | 1 (10.0%) | 0 (0.0%) | 7 (70.0%) | 0 (0.0%) | 7 (70.0%) | 7 (70.0%) | 8 (80.0%) |
| sample_930 | 32 | 853 | 27 (84.4%) | 4 (12.5%) | 1 (3.1%) | 0 (0.0%) | 6 (18.8%) | 1 (3.1%) | 7 (21.9%) | 11 (34.4%) |
| sample_942 | 10 | 285 | 8 (80.0%) | 0 (0.0%) | 0 (0.0%) | 2 (20.0%) | 0 (0.0%) | 2 (20.0%) | 2 (20.0%) | 2 (20.0%) |
| sample_945 | 24 | 766 | 16 (66.7%) | 3 (12.5%) | 1 (4.2%) | 4 (16.7%) | 7 (29.2%) | 5 (20.8%) | 12 (50.0%) | 15 (62.5%) |
| sample_96 | 23 | 759 | 12 (52.2%) | 5 (21.7%) | 0 (0.0%) | 7 (30.4%) | 4 (17.4%) | 7 (30.4%) | 11 (47.8%) | 16 (69.6%) |
| Correlation on counts | | | 0.68342 | 0.30926 | 0.26566 | 0.29055 | 0.51050 | 0.34004 | 0.63495 | 0.64637 |
| p-value on counts | | | 0.00000 | 0.01530 | 0.03852 | 0.02311 | 0.00003 | 0.00734 | 0.00000 | 0.00000 |
| Correlation on rates | | | 0.27939 | -0.07564 | -0.10315 | -0.20635 | 0.10905 | -0.24225 | 0.02410 | 0.00783 |
| p-value on rates | | | 0.02921 | 0.56230 | 0.42890 | 0.11060 | 0.40280 | 0.05997 | 0.85380 | 0.95220 |
| GatorTronS + CLAMP | | | | | | | | | | |
| Sample | Anno # | Word # | EM | RM | MM | UD | OD | MMUD | Error | RM+Error |
| sample_1128 | 33 | 492 | 16 (48.5%) | 1 (3.0%) | 1 (3.0%) | 15 (45.5%) | 0 (0.0%) | 16 (48.5%) | 16 (48.5%) | 17 (51.5%) |
| sample_1133 | 24 | 782 | 10 (41.7%) | 1 (4.2%) | 1 (4.2%) | 12 (50.0%) | 1 (4.2%) | 13 (54.2%) | 14 (58.3%) | 15 (62.5%) |
| sample_1152 | 18 | 593 | 10 (55.6%) | 1 (5.6%) | 2 (11.1%) | 5 (27.8%) | 2 (11.1%) | 7 (38.9%) | 9 (50.0%) | 10 (55.6%) |
| sample_1169 | 8 | 422 | 3 (37.5%) | 2 (25.0%) | 1 (12.5%) | 3 (37.5%) | 0 (0.0%) | 4 (50.0%) | 4 (50.0%) | 6 (75.0%) |
| sample_1242 | 11 | 536 | 3 (27.3%) | 0 (0.0%) | 2 (18.2%) | 6 (54.5%) | 4 (36.4%) | 8 (72.7%) | 12 (109.1%) | 12 (109.1%) |
| sample_1248 | 31 | 910 | 13 (41.9%) | 0 (0.0%) | 5 (16.1%) | 13 (41.9%) | 10 (32.3%) | 18 (58.1%) | 28 (90.3%) | 28 (90.3%) |

| sample | | | | | | | | | | |
|---|---|---|---|---|---|---|---|---|---|---|
| sample_1252 | 19 | 585 | 10 (52.6%) | 4 (21.1%) | 0 (0.0%) | 6 (31.6%) | 1 (5.3%) | 6 (31.6%) | 7 (36.8%) | 11 (57.9%) |
| sample_1419 | 10 | 540 | 2 (20.0%) | 5 (50.0%) | 2 (20.0%) | 2 (20.0%) | 1 (10.0%) | 4 (40.0%) | 5 (50.0%) | 10 (100.0%) |
| sample_1439 | 19 | 896 | 4 (21.1%) | 4 (21.1%) | 2 (10.5%) | 9 (47.4%) | 3 (15.8%) | 11 (57.9%) | 14 (73.7%) | 18 (94.7%) |
| sample_1495 | 45 | 872 | 17 (37.8%) | 1 (2.2%) | 4 (8.9%) | 23 (51.1%) | 1 (2.2%) | 27 (60.0%) | 28 (62.2%) | 29 (64.4%) |
| sample_1505 | 16 | 784 | 6 (37.5%) | 3 (18.8%) | 2 (12.5%) | 5 (31.3%) | 7 (43.8%) | 7 (43.8%) | 14 (87.5%) | 17 (106.3%) |
| sample_1568 | 28 | 1112 | 19 (67.9%) | 3 (10.7%) | 2 (7.1%) | 4 (14.3%) | 11 (39.3%) | 6 (21.4%) | 17 (60.7%) | 20 (71.4%) |
| sample_1592 | 24 | 785 | 16 (66.7%) | 2 (8.3%) | 0 (0.0%) | 6 (25.0%) | 3 (12.5%) | 6 (25.0%) | 9 (37.5%) | 11 (45.8%) |
| sample_1921 | 21 | 789 | 12 (57.1%) | 3 (14.3%) | 1 (4.8%) | 6 (28.6%) | 3 (14.3%) | 7 (33.3%) | 10 (47.6%) | 13 (61.9%) |
| sample_1956 | 14 | 628 | 5 (35.7%) | 1 (7.1%) | 1 (7.1%) | 7 (50.0%) | 10 (71.4%) | 8 (57.1%) | 18 (128.6%) | 19 (135.7%) |
| sample_2129 | 13 | 427 | 3 (23.1%) | 5 (38.5%) | 2 (15.4%) | 4 (30.8%) | 3 (23.1%) | 6 (46.2%) | 9 (69.2%) | 14 (107.7%) |
| sample_214 | 14 | 586 | 4 (28.6%) | 2 (14.3%) | 3 (21.4%) | 5 (35.7%) | 4 (28.6%) | 8 (57.1%) | 12 (85.7%) | 14 (100.0%) |
| sample_2210 | 38 | 874 | 5 (13.2%) | 3 (7.9%) | 0 (0.0%) | 30 (78.9%) | 0 (0.0%) | 30 (78.9%) | 30 (78.9%) | 33 (86.8%) |
| sample_2218 | 35 | 771 | 16 (45.7%) | 8 (22.9%) | 2 (5.7%) | 9 (25.7%) | 3 (8.6%) | 11 (31.4%) | 14 (40.0%) | 22 (62.9%) |
| sample_223 | 11 | 567 | 3 (27.3%) | 1 (9.1%) | 2 (18.2%) | 5 (45.5%) | 0 (0.0%) | 7 (63.6%) | 7 (63.6%) | 8 (72.7%) |
| sample_225 | 13 | 468 | 4 (30.8%) | 2 (15.4%) | 0 (0.0%) | 7 (53.8%) | 0 (0.0%) | 7 (53.8%) | 7 (53.8%) | 9 (69.2%) |
| sample_226 | 8 | 1022 | 1 (12.5%) | 2 (25.0%) | 1 (12.5%) | 4 (50.0%) | 10 (125.0%) | 5 (62.5%) | 15 (187.5%) | 17 (212.5%) |
| sample_2275 | 27 | 612 | 9 (33.3%) | 1 (3.7%) | 3 (11.1%) | 14 (51.9%) | 2 (7.4%) | 17 (63.0%) | 19 (70.4%) | 20 (74.1%) |
| sample_2604 | 20 | 733 | 4 (20.0%) | 2 (10.0%) | 0 (0.0%) | 15 (75.0%) | 1 (5.0%) | 15 (75.0%) | 16 (80.0%) | 18 (90.0%) |
| sample_2623 | 14 | 480 | 6 (42.9%) | 0 (0.0%) | 3 (21.4%) | 5 (35.7%) | 6 (42.9%) | 8 (57.1%) | 14 (100.0%) | 14 (100.0%) |
| sample_2746 | 31 | 2066 | 23 (74.2%) | 6 (19.4%) | 1 (3.2%) | 2 (6.5%) | 17 (54.8%) | 3 (9.7%) | 20 (64.5%) | 26 (83.9%) |
| sample_2747 | 13 | 814 | 5 (38.5%) | 1 (7.7%) | 0 (0.0%) | 7 (53.8%) | 2 (15.4%) | 7 (53.8%) | 9 (69.2%) | 10 (76.9%) |
| sample_2780 | 48 | 994 | 21 (43.8%) | 11 (22.9%) | 4 (8.3%) | 14 (29.2%) | 6 (12.5%) | 18 (37.5%) | 24 (50.0%) | 35 (72.9%) |
| sample_2789 | 34 | 1082 | 23 (67.6%) | 6 (17.6%) | 1 (2.9%) | 5 (14.7%) | 13 (38.2%) | 6 (17.6%) | 19 (55.9%) | 25 (73.5%) |
| sample_2790 | 34 | 1123 | 22 (64.7%) | 8 (23.5%) | 1 (2.9%) | 5 (14.7%) | 19 (55.9%) | 6 (17.6%) | 25 (73.5%) | 33 (97.1%) |
| sample_2792 | 9 | 708 | 5 (55.6%) | 1 (11.1%) | 1 (11.1%) | 2 (22.2%) | 28 (311.1%) | 3 (33.3%) | 31 (344.4%) | 32 (355.6%) |
| sample_343 | 25 | 1075 | 12 (48.0%) | 6 (24.0%) | 3 (12.0%) | 4 (16.0%) | 8 (32.0%) | 7 (28.0%) | 15 (60.0%) | 21 (84.0%) |
| sample_365 | 57 | 1312 | 39 (68.4%) | 9 (15.8%) | 6 (10.5%) | 4 (7.0%) | 8 (14.0%) | 10 (17.5%) | 18 (31.6%) | 27 (47.4%) |
| sample_377 | 55 | 1420 | 21 (38.2%) | 8 (14.5%) | 8 (14.5%) | 19 (34.5%) | 14 (25.5%) | 27 (49.1%) | 41 (74.5%) | 49 (89.1%) |
| sample_378 | 24 | 939 | 11 (45.8%) | 4 (16.7%) | 2 (8.3%) | 8 (33.3%) | 7 (29.2%) | 10 (41.7%) | 17 (70.8%) | 21 (87.5%) |
| sample_380 | 50 | 1139 | 21 (42.0%) | 3 (6.0%) | 1 (2.0%) | 26 (52.0%) | 4 (8.0%) | 27 (54.0%) | 31 (62.0%) | 34 (68.0%) |
| sample_388 | 17 | 620 | 5 (29.4%) | 3 (17.6%) | 3 (17.6%) | 6 (35.3%) | 12 (70.6%) | 9 (52.9%) | 21 (123.5%) | 24 (141.2%) |

| Sample | Anno # | Word # | EM | RM | MM | UD | OD | MMUD | Error | RM+Error |
|---|---|---|---|---|---|---|---|---|---|---|
| sample_391 | 20 | 918 | 2 (10.0%) | 4 (20.0%) | 4 (20.0%) | 10 (50.0%) | 6 (30.0%) | 14 (70.0%) | 20 (100.0%) | 24 (120.0%) |
| sample_392 | 34 | 1004 | 13 (38.2%) | 2 (5.9%) | 2 (5.9%) | 17 (50.0%) | 2 (5.9%) | 19 (55.9%) | 21 (61.8%) | 23 (67.6%) |
| sample_393 | 50 | 1552 | 31 (62.0%) | 6 (12.0%) | 1 (2.0%) | 14 (28.0%) | 9 (18.0%) | 15 (30.0%) | 24 (48.0%) | 30 (60.0%) |
| sample_394 | 19 | 676 | 8 (42.1%) | 0 (0.0%) | 4 (21.1%) | 7 (36.8%) | 3 (15.8%) | 11 (57.9%) | 14 (73.7%) | 14 (73.7%) |
| sample_398 | 24 | 694 | 18 (75.0%) | 4 (16.7%) | 0 (0.0%) | 2 (8.3%) | 3 (12.5%) | 2 (8.3%) | 5 (20.8%) | 9 (37.5%) |
| sample_402 | 15 | 491 | 9 (60.0%) | 2 (13.3%) | 3 (20.0%) | 1 (6.7%) | 2 (13.3%) | 4 (26.7%) | 6 (40.0%) | 8 (53.3%) |
| sample_403 | 17 | 562 | 11 (64.7%) | 4 (23.5%) | 0 (0.0%) | 2 (11.8%) | 3 (17.6%) | 2 (11.8%) | 5 (29.4%) | 9 (52.9%) |
| sample_439 | 16 | 411 | 5 (31.3%) | 2 (12.5%) | 2 (12.5%) | 7 (43.8%) | 1 (6.3%) | 9 (56.3%) | 10 (62.5%) | 12 (75.0%) |
| sample_452 | 15 | 398 | 9 (60.0%) | 1 (6.7%) | 1 (6.7%) | 4 (26.7%) | 1 (6.7%) | 5 (33.3%) | 6 (40.0%) | 7 (46.7%) |
| sample_476 | 17 | 666 | 6 (35.3%) | 6 (35.3%) | 2 (11.8%) | 3 (17.6%) | 4 (23.5%) | 5 (29.4%) | 9 (52.9%) | 15 (88.2%) |
| sample_485 | 24 | 742 | 10 (41.7%) | 2 (8.3%) | 2 (8.3%) | 10 (41.7%) | 4 (16.7%) | 12 (50.0%) | 16 (66.7%) | 18 (75.0%) |
| sample_570 | 31 | 1038 | 17 (54.8%) | 5 (16.1%) | 5 (16.1%) | 5 (16.1%) | 15 (48.4%) | 10 (32.3%) | 25 (80.6%) | 30 (96.8%) |
| sample_579 | 30 | 932 | 16 (53.3%) | 7 (23.3%) | 2 (6.7%) | 6 (20.0%) | 3 (10.0%) | 8 (26.7%) | 11 (36.7%) | 18 (60.0%) |
| sample_583 | 29 | 927 | 18 (62.1%) | 5 (17.2%) | 0 (0.0%) | 6 (20.7%) | 12 (41.4%) | 6 (20.7%) | 18 (62.1%) | 23 (79.3%) |
| sample_664 | 19 | 885 | 12 (63.2%) | 2 (10.5%) | 2 (10.5%) | 3 (15.8%) | 2 (10.5%) | 5 (26.3%) | 7 (36.8%) | 9 (47.4%) |
| sample_666 | 40 | 873 | 23 (57.5%) | 7 (17.5%) | 2 (5.0%) | 11 (27.5%) | 5 (12.5%) | 13 (32.5%) | 18 (45.0%) | 25 (62.5%) |
| sample_687 | 20 | 426 | 8 (40.0%) | 6 (30.0%) | 3 (15.0%) | 5 (25.0%) | 0 (0.0%) | 8 (40.0%) | 8 (40.0%) | 14 (70.0%) |
| sample_70 | 7 | 519 | 5 (71.4%) | 1 (14.3%) | 0 (0.0%) | 1 (14.3%) | 1 (14.3%) | 1 (14.3%) | 2 (28.6%) | 3 (42.9%) |
| sample_71 | 12 | 562 | 6 (50.0%) | 1 (8.3%) | 0 (0.0%) | 5 (41.7%) | 2 (16.7%) | 5 (41.7%) | 7 (58.3%) | 8 (66.7%) |
| sample_782 | 10 | 355 | 3 (30.0%) | 1 (10.0%) | 0 (0.0%) | 6 (60.0%) | 0 (0.0%) | 6 (60.0%) | 6 (60.0%) | 7 (70.0%) |
| sample_930 | 32 | 853 | 24 (75.0%) | 3 (9.4%) | 3 (9.4%) | 2 (6.3%) | 7 (21.9%) | 5 (15.6%) | 12 (37.5%) | 15 (46.9%) |
| sample_942 | 10 | 285 | 8 (80.0%) | 1 (10.0%) | 1 (10.0%) | 0 (0.0%) | 1 (10.0%) | 1 (10.0%) | 2 (20.0%) | 3 (30.0%) |
| sample_945 | 24 | 766 | 14 (58.3%) | 3 (12.5%) | 2 (8.3%) | 5 (20.8%) | 10 (41.7%) | 7 (29.2%) | 17 (70.8%) | 20 (83.3%) |
| sample_96 | 23 | 759 | 13 (56.5%) | 6 (26.1%) | 0 (0.0%) | 5 (21.7%) | 3 (13.0%) | 5 (21.7%) | 8 (34.8%) | 14 (60.9%) |
| Correlation on counts | | | 0.68036 | 0.51942 | 0.26845 | 0.26135 | 0.56099 | 0.31017 | 0.62392 | 0.70026 |
| p-value on counts | | | 0.00000 | 0.00002 | 0.03645 | 0.04190 | 0.00000 | 0.01499 | 0.00000 | 0.00000 |
| Correlation on rates | | | 0.23659 | 0.02897 | -0.17122 | -0.18013 | 0.13932 | -0.23490 | 0.03774 | 0.04391 |
| p-value on rates | | | 0.06640 | 0.82460 | 0.18710 | 0.16480 | 0.28430 | 0.06842 | 0.77280 | 0.73680 |
| GatorTron | | | | | | | | | | |
| Sample | Anno # | Word # | EM | RM | MM | UD | OD | MMUD | Error | RM+Error |
| sample_1128 | 33 | 492 | 18 (54.5%) | 1 (3.0%) | 2 (6.1%) | 12 (36.4%) | 2 (6.1%) | 14 (42.4%) | 16 (48.5%) | 17 (51.5%) |
| sample_1133 | 24 | 782 | 11 (45.8%) | 2 (8.3%) | 1 (4.2%) | 10 (41.7%) | 5 (20.8%) | 11 (45.8%) | 16 (66.7%) | 18 (75.0%) |
| sample_1152 | 18 | 593 | 11 (61.1%) | 1 (5.6%) | 3 (16.7%) | 3 (16.7%) | 1 (5.6%) | 6 (33.3%) | 7 (38.9%) | 8 (44.4%) |
| sample_1169 | 8 | 422 | 4 (50.0%) | 0 (0.0%) | 1 (12.5%) | 3 (37.5%) | 0 (0.0%) | 4 (50.0%) | 4 (50.0%) | 4 (50.0%) |
| sample_1242 | 11 | 536 | 3 (27.3%) | 1 (9.1%) | 1 (9.1%) | 6 (54.5%) | 5 (45.5%) | 7 (63.6%) | 12 (109.1%) | 13 (118.2%) |
| sample_1248 | 31 | 910 | 9 (29.0%) | 5 (16.1%) | 7 (22.6%) | 10 (32.3%) | 16 (51.6%) | 17 (54.8%) | 33 (106.5%) | 38 (122.6%) |
| sample_1252 | 19 | 585 | 9 (47.4%) | 7 (36.8%) | 1 (5.3%) | 3 (15.8%) | 1 (5.3%) | 4 (21.1%) | 5 (26.3%) | 12 (63.2%) |
| sample_1419 | 10 | 540 | 3 (30.0%) | 4 (40.0%) | 2 (20.0%) | 2 (20.0%) | 1 (10.0%) | 4 (40.0%) | 5 (50.0%) | 9 (90.0%) |
| sample_1439 | 19 | 896 | 4 (21.1%) | 3 (15.8%) | 4 (21.1%) | 8 (42.1%) | 4 (21.1%) | 12 (63.2%) | 16 (84.2%) | 19 (100.0%) |
| sample_1495 | 45 | 872 | 18 (40.0%) | 0 (0.0%) | 4 (8.9%) | 23 (51.1%) | 2 (4.4%) | 27 (60.0%) | 29 (64.4%) | 29 (64.4%) |
| sample_1505 | 16 | 784 | 7 (43.8%) | 3 (18.8%) | 4 (25.0%) | 2 (12.5%) | 7 (43.8%) | 6 (37.5%) | 13 (81.3%) | 16 (100.0%) |
| sample_1568 | 28 | 1112 | 21 (75.0%) | 2 (7.1%) | 2 (7.1%) | 3 (10.7%) | 15 (53.6%) | 5 (17.9%) | 20 (71.4%) | 22 (78.6%) |
| sample_1592 | 24 | 785 | 19 (79.2%) | 2 (8.3%) | 0 (0.0%) | 3 (12.5%) | 3 (12.5%) | 3 (12.5%) | 6 (25.0%) | 8 (33.3%) |
| sample_1921 | 21 | 789 | 12 (57.1%) | 5 (23.8%) | 2 (9.5%) | 2 (9.5%) | 10 (47.6%) | 4 (19.0%) | 14 (66.7%) | 19 (90.5%) |
| sample_1956 | 14 | 628 | 9 (64.3%) | 2 (14.3%) | 0 (0.0%) | 3 (21.4%) | 14 (100.0%) | 3 (21.4%) | 17 (121.4%) | 19 (135.7%) |
| sample_2129 | 13 | 427 | 3 (23.1%) | 5 (38.5%) | 1 (7.7%) | 4 (30.8%) | 12 (92.3%) | 5 (38.5%) | 17 (130.8%) | 22 (169.2%) |
| sample_214 | 14 | 586 | 7 (50.0%) | 2 (14.3%) | 5 (35.7%) | 0 (0.0%) | 5 (35.7%) | 5 (35.7%) | 10 (71.4%) | 12 (85.7%) |
| sample_2210 | 38 | 874 | 9 (23.7%) | 2 (5.3%) | 1 (2.6%) | 26 (68.4%) | 2 (5.3%) | 27 (71.1%) | 29 (76.3%) | 31 (81.6%) |
| sample_2218 | 35 | 771 | 21 (60.0%) | 8 (22.9%) | 4 (11.4%) | 3 (8.6%) | 5 (14.3%) | 7 (20.0%) | 12 (34.3%) | 20 (57.1%) |
| sample_223 | 11 | 567 | 4 (36.4%) | 5 (45.5%) | 2 (18.2%) | 2 (18.2%) | 4 (36.4%) | 4 (36.4%) | 8 (72.7%) | 13 (118.2%) |
| sample_225 | 13 | 468 | 5 (38.5%) | 7 (53.8%) | 1 (7.7%) | 2 (15.4%) | 2 (15.4%) | 3 (23.1%) | 5 (38.5%) | 12 (92.3%) |
| sample_226 | 8 | 1022 | 3 (37.5%) | 1 (12.5%) | 2 (25.0%) | 2 (25.0%) | 13 (162.5%) | 4 (50.0%) | 17 (212.5%) | 18 (225.0%) |
| sample_2275 | 27 | 612 | 12 (44.4%) | 4 (14.8%) | 2 (7.4%) | 9 (33.3%) | 3 (11.1%) | 11 (40.7%) | 14 (51.9%) | 18 (66.7%) |
| sample_2604 | 20 | 733 | 5 (25.0%) | 3 (15.0%) | 4 (20.0%) | 9 (45.0%) | 6 (30.0%) | 13 (65.0%) | 19 (95.0%) | 22 (110.0%) |
| sample_2623 | 14 | 480 | 9 (64.3%) | 0 (0.0%) | 3 (21.4%) | 2 (14.3%) | 7 (50.0%) | 5 (35.7%) | 12 (85.7%) | 12 (85.7%) |
| sample_2746 | 31 | 2066 | 23 (74.2%) | 7 (22.6%) | 0 (0.0%) | 2 (6.5%) | 19 (61.3%) | 2 (6.5%) | 21 (67.7%) | 28 (90.3%) |
| sample_2747 | 13 | 814 | 4 (30.8%) | 0 (0.0%) | 0 (0.0%) | 9 (69.2%) | 1 (7.7%) | 9 (69.2%) | 10 (76.9%) | 10 (76.9%) |
| sample_2780 | 48 | 994 | 23 (47.9%) | 12 (25.0%) | 4 (8.3%) | 14 (29.2%) | 4 (8.3%) | 18 (37.5%) | 22 (45.8%) | 34 (70.8%) |
| sample_2789 | 34 | 1082 | 25 (73.5%) | 5 (14.7%) | 2 (5.9%) | 4 (11.8%) | 16 (47.1%) | 6 (17.6%) | 22 (64.7%) | 27 (79.4%) |
| sample_2790 | 34 | 1123 | 22 (64.7%) | 4 (11.8%) | 2 (5.9%) | 6 (17.6%) | 21 (61.8%) | 8 (23.5%) | 29 (85.3%) | 33 (97.1%) |
| sample_2792 | 9 | 708 | 4 (44.4%) | 1 (11.1%) | 2 (22.2%) | 2 (22.2%) | 26 (288.9%) | 4 (44.4%) | 30 (333.3%) | 31 (344.4%) |
| sample_343 | 25 | 1075 | 13 (52.0%) | 8 (32.0%) | 3 (12.0%) | 3 (12.0%) | 10 (40.0%) | 6 (24.0%) | 16 (64.0%) | 24 (96.0%) |

| sample | | | | | | | | | | |
|---|---|---|---|---|---|---|---|---|---|---|
| sample_365 | 57 | 1312 | 39 (68.4%) | 12 (21.1%) | 5 (8.8%) | 3 (5.3%) | 16 (28.1%) | 8 (14.0%) | 24 (42.1%) | 36 (63.2%) |
| sample_377 | 55 | 1420 | 27 (49.1%) | 7 (12.7%) | 10 (18.2%) | 11 (20.0%) | 10 (18.2%) | 21 (38.2%) | 31 (56.4%) | 38 (69.1%) |
| sample_378 | 24 | 939 | 10 (41.7%) | 3 (12.5%) | 1 (4.2%) | 10 (41.7%) | 3 (12.5%) | 11 (45.8%) | 14 (58.3%) | 17 (70.8%) |
| sample_380 | 50 | 1139 | 23 (46.0%) | 2 (4.0%) | 3 (6.0%) | 22 (44.0%) | 2 (4.0%) | 25 (50.0%) | 27 (54.0%) | 29 (58.0%) |
| sample_388 | 17 | 620 | 7 (41.2%) | 3 (17.6%) | 4 (23.5%) | 3 (17.6%) | 9 (52.9%) | 7 (41.2%) | 16 (94.1%) | 19 (111.8%) |
| sample_391 | 20 | 918 | 4 (20.0%) | 4 (20.0%) | 6 (30.0%) | 6 (30.0%) | 8 (40.0%) | 12 (60.0%) | 20 (100.0%) | 24 (120.0%) |
| sample_392 | 34 | 1004 | 14 (41.2%) | 7 (20.6%) | 2 (5.9%) | 12 (35.3%) | 5 (14.7%) | 14 (41.2%) | 19 (55.9%) | 26 (76.5%) |
| sample_393 | 50 | 1552 | 32 (64.0%) | 2 (4.0%) | 3 (6.0%) | 13 (26.0%) | 10 (20.0%) | 16 (32.0%) | 26 (52.0%) | 28 (56.0%) |
| sample_394 | 19 | 676 | 6 (31.6%) | 1 (5.3%) | 4 (21.1%) | 8 (42.1%) | 2 (10.5%) | 12 (63.2%) | 14 (73.7%) | 15 (78.9%) |
| sample_398 | 24 | 694 | 17 (70.8%) | 2 (8.3%) | 1 (4.2%) | 4 (16.7%) | 2 (8.3%) | 5 (20.8%) | 7 (29.2%) | 9 (37.5%) |
| sample_402 | 15 | 491 | 9 (60.0%) | 3 (20.0%) | 2 (13.3%) | 1 (6.7%) | 2 (13.3%) | 3 (20.0%) | 5 (33.3%) | 8 (53.3%) |
| sample_403 | 17 | 562 | 9 (52.9%) | 6 (35.3%) | 3 (17.6%) | 0 (0.0%) | 2 (11.8%) | 3 (17.6%) | 5 (29.4%) | 11 (64.7%) |
| sample_439 | 16 | 411 | 7 (43.8%) | 1 (6.3%) | 3 (18.8%) | 5 (31.3%) | 2 (12.5%) | 8 (50.0%) | 10 (62.5%) | 11 (68.8%) |
| sample_452 | 15 | 398 | 8 (53.3%) | 4 (26.7%) | 2 (13.3%) | 2 (13.3%) | 2 (13.3%) | 4 (26.7%) | 6 (40.0%) | 10 (66.7%) |
| sample_476 | 17 | 666 | 5 (29.4%) | 7 (41.2%) | 2 (11.8%) | 3 (17.6%) | 5 (29.4%) | 5 (29.4%) | 10 (58.8%) | 17 (100.0%) |
| sample_485 | 24 | 742 | 7 (29.2%) | 6 (25.0%) | 3 (12.5%) | 10 (41.7%) | 6 (25.0%) | 13 (54.2%) | 19 (79.2%) | 25 (104.2%) |
| sample_570 | 31 | 1038 | 15 (48.4%) | 11 (35.5%) | 3 (9.7%) | 3 (9.7%) | 14 (45.2%) | 6 (19.4%) | 20 (64.5%) | 31 (100.0%) |
| sample_579 | 30 | 932 | 20 (66.7%) | 2 (6.7%) | 4 (13.3%) | 4 (13.3%) | 6 (20.0%) | 8 (26.7%) | 14 (46.7%) | 16 (53.3%) |
| sample_583 | 29 | 927 | 17 (58.6%) | 3 (10.3%) | 1 (3.4%) | 8 (27.6%) | 12 (41.4%) | 9 (31.0%) | 21 (72.4%) | 24 (82.8%) |
| sample_664 | 19 | 885 | 12 (63.2%) | 0 (0.0%) | 3 (15.8%) | 4 (21.1%) | 3 (15.8%) | 7 (36.8%) | 10 (52.6%) | 10 (52.6%) |
| sample_666 | 40 | 873 | 28 (70.0%) | 4 (10.0%) | 1 (2.5%) | 9 (22.5%) | 6 (15.0%) | 10 (25.0%) | 16 (40.0%) | 20 (50.0%) |
| sample_687 | 20 | 426 | 8 (40.0%) | 8 (40.0%) | 0 (0.0%) | 7 (35.0%) | 0 (0.0%) | 7 (35.0%) | 7 (35.0%) | 15 (75.0%) |
| sample_70 | 7 | 519 | 4 (57.1%) | 0 (0.0%) | 0 (0.0%) | 3 (42.9%) | 1 (14.3%) | 3 (42.9%) | 4 (57.1%) | 4 (57.1%) |
| sample_71 | 12 | 562 | 2 (16.7%) | 1 (8.3%) | 0 (0.0%) | 9 (75.0%) | 2 (16.7%) | 9 (75.0%) | 11 (91.7%) | 12 (100.0%) |
| sample_782 | 10 | 355 | 4 (40.0%) | 2 (20.0%) | 0 (0.0%) | 4 (40.0%) | 0 (0.0%) | 4 (40.0%) | 4 (40.0%) | 6 (60.0%) |
| sample_930 | 32 | 853 | 26 (81.3%) | 3 (9.4%) | 2 (6.3%) | 1 (3.1%) | 10 (31.3%) | 3 (9.4%) | 13 (40.6%) | 16 (50.0%) |
| sample_942 | 10 | 285 | 8 (80.0%) | 1 (10.0%) | 1 (10.0%) | 0 (0.0%) | 1 (10.0%) | 1 (10.0%) | 2 (20.0%) | 3 (30.0%) |
| sample_945 | 24 | 766 | 15 (62.5%) | 5 (20.8%) | 1 (4.2%) | 3 (12.5%) | 7 (29.2%) | 4 (16.7%) | 11 (45.8%) | 16 (66.7%) |
| sample_96 | 23 | 759 | 13 (56.5%) | 7 (30.4%) | 0 (0.0%) | 4 (17.4%) | 5 (21.7%) | 4 (17.4%) | 9 (39.1%) | 16 (69.6%) |
| Correlation on counts | | | 0.67188 | 0.31818 | 0.29032 | 0.26406 | 0.56324 | 0.33014 | 0.66493 | 0.69897 |
| p-value on counts | | | 0.00000 | 0.01246 | 0.02323 | 0.03975 | 0.00000 | 0.00937 | 0.00000 | 0.00000 |
| Correlation on rates | | | 0.25202 | -0.13217 | -0.11758 | -0.11359 | 0.12332 | -0.17145 | 0.05163 | 0.01489 |
| p-value on rates | | | 0.05007 | 0.30990 | 0.36680 | 0.38340 | 0.34370 | 0.18650 | 0.69270 | 0.90930 |
| GatorTron + CLAMP | | | | | | | | | | |
| Sample | Anno # | Word # | EM | RM | MM | UD | OD | MMUD | Error | RM+Error |
| sample_1128 | 33 | 492 | 21 (63.6%) | 3 (9.1%) | 1 (3.0%) | 8 (24.2%) | 3 (9.1%) | 9 (27.3%) | 12 (36.4%) | 15 (45.5%) |
| sample_1133 | 24 | 782 | 12 (50.0%) | 2 (8.3%) | 1 (4.2%) | 9 (37.5%) | 6 (25.0%) | 10 (41.7%) | 16 (66.7%) | 18 (75.0%) |
| sample_1152 | 18 | 593 | 8 (44.4%) | 3 (16.7%) | 3 (16.7%) | 4 (22.2%) | 3 (16.7%) | 7 (38.9%) | 10 (55.6%) | 13 (72.2%) |
| sample_1169 | 8 | 422 | 4 (50.0%) | 0 (0.0%) | 2 (25.0%) | 2 (25.0%) | 1 (12.5%) | 4 (50.0%) | 5 (62.5%) | 5 (62.5%) |
| sample_1242 | 11 | 536 | 3 (27.3%) | 2 (18.2%) | 1 (9.1%) | 6 (54.5%) | 3 (27.3%) | 7 (63.6%) | 10 (90.9%) | 12 (109.1%) |
| sample_1248 | 31 | 910 | 10 (32.3%) | 5 (16.1%) | 8 (25.8%) | 9 (29.0%) | 19 (61.3%) | 17 (54.8%) | 36 (116.1%) | 41 (132.3%) |
| sample_1252 | 19 | 585 | 9 (47.4%) | 7 (36.8%) | 1 (5.3%) | 3 (15.8%) | 1 (5.3%) | 4 (21.1%) | 5 (26.3%) | 12 (63.2%) |
| sample_1419 | 10 | 540 | 5 (50.0%) | 4 (40.0%) | 2 (20.0%) | 0 (0.0%) | 2 (20.0%) | 2 (20.0%) | 4 (40.0%) | 8 (80.0%) |
| sample_1439 | 19 | 896 | 5 (26.3%) | 6 (31.6%) | 4 (21.1%) | 5 (26.3%) | 6 (31.6%) | 9 (47.4%) | 15 (78.9%) | 21 (110.5%) |
| sample_1495 | 45 | 872 | 17 (37.8%) | 2 (4.4%) | 6 (13.3%) | 20 (44.4%) | 5 (11.1%) | 26 (57.8%) | 31 (68.9%) | 33 (73.3%) |
| sample_1505 | 16 | 784 | 7 (43.8%) | 3 (18.8%) | 4 (25.0%) | 2 (12.5%) | 7 (43.8%) | 6 (37.5%) | 13 (81.3%) | 16 (100.0%) |
| sample_1568 | 28 | 1112 | 20 (71.4%) | 4 (14.3%) | 2 (7.1%) | 3 (10.7%) | 18 (64.3%) | 5 (17.9%) | 23 (82.1%) | 27 (96.4%) |
| sample_1592 | 24 | 785 | 18 (75.0%) | 2 (8.3%) | 0 (0.0%) | 4 (16.7%) | 3 (12.5%) | 4 (16.7%) | 7 (29.2%) | 9 (37.5%) |
| sample_1921 | 21 | 789 | 12 (57.1%) | 3 (14.3%) | 1 (4.8%) | 5 (23.8%) | 6 (28.6%) | 6 (28.6%) | 12 (57.1%) | 15 (71.4%) |
| sample_1956 | 14 | 628 | 4 (28.6%) | 0 (0.0%) | 1 (7.1%) | 9 (64.3%) | 6 (42.9%) | 10 (71.4%) | 16 (114.3%) | 16 (114.3%) |
| sample_2129 | 13 | 427 | 3 (23.1%) | 4 (30.8%) | 2 (15.4%) | 4 (30.8%) | 10 (76.9%) | 6 (46.2%) | 16 (123.1%) | 20 (153.8%) |
| sample_214 | 14 | 586 | 3 (21.4%) | 4 (28.6%) | 5 (35.7%) | 2 (14.3%) | 3 (21.4%) | 7 (50.0%) | 10 (71.4%) | 14 (100.0%) |
| sample_2210 | 38 | 874 | 8 (21.1%) | 2 (5.3%) | 0 (0.0%) | 28 (73.7%) | 2 (5.3%) | 28 (73.7%) | 30 (78.9%) | 32 (84.2%) |
| sample_2218 | 35 | 771 | 18 (51.4%) | 6 (17.1%) | 2 (5.7%) | 9 (25.7%) | 4 (11.4%) | 11 (31.4%) | 15 (42.9%) | 21 (60.0%) |
| sample_223 | 11 | 567 | 4 (36.4%) | 3 (27.3%) | 1 (9.1%) | 4 (36.4%) | 2 (18.2%) | 5 (45.5%) | 7 (63.6%) | 10 (90.9%) |
| sample_225 | 13 | 468 | 4 (30.8%) | 8 (61.5%) | 0 (0.0%) | 5 (38.5%) | 1 (7.7%) | 5 (38.5%) | 6 (46.2%) | 14 (107.7%) |
| sample_226 | 8 | 1022 | 2 (25.0%) | 2 (25.0%) | 2 (25.0%) | 2 (25.0%) | 10 (125.0%) | 4 (50.0%) | 14 (175.0%) | 16 (200.0%) |
| sample_2275 | 27 | 612 | 12 (44.4%) | 2 (7.4%) | 2 (7.4%) | 11 (40.7%) | 3 (11.1%) | 13 (48.1%) | 16 (59.3%) | 18 (66.7%) |
| sample_2604 | 20 | 733 | 3 (15.0%) | 4 (20.0%) | 0 (0.0%) | 13 (65.0%) | 1 (5.0%) | 13 (65.0%) | 14 (70.0%) | 18 (90.0%) |
| sample_2623 | 14 | 480 | 8 (57.1%) | 0 (0.0%) | 2 (14.3%) | 4 (28.6%) | 3 (21.4%) | 6 (42.9%) | 9 (64.3%) | 9 (64.3%) |
| sample_2746 | 31 | 2066 | 23 (74.2%) | 6 (19.4%) | 2 (6.5%) | 1 (3.2%) | 18 (58.1%) | 3 (9.7%) | 21 (67.7%) | 27 (87.1%) |
| sample_2747 | 13 | 814 | 5 (38.5%) | 0 (0.0%) | 1 (7.7%) | 7 (53.8%) | 0 (0.0%) | 8 (61.5%) | 8 (61.5%) | 8 (61.5%) |

| | | | | | | | | | |
|---|---|---|---|---|---|---|---|---|---|
| sample_2780 | 48 | 994 | 22 (45.8%) | 9 (18.8%) | 4 (8.3%) | 15 (31.3%) | 5 (10.4%) | 19 (39.6%) | 24 (50.0%) | 33 (68.8%) |
| sample_2789 | 34 | 1082 | 24 (70.6%) | 6 (17.6%) | 2 (5.9%) | 4 (11.8%) | 14 (41.2%) | 6 (17.6%) | 20 (58.8%) | 26 (76.5%) |
| sample_2790 | 34 | 1123 | 20 (58.8%) | 8 (23.5%) | 1 (2.9%) | 7 (20.6%) | 20 (58.8%) | 8 (23.5%) | 28 (82.4%) | 36 (105.9%) |
| sample_2792 | 9 | 708 | 4 (44.4%) | 2 (22.2%) | 1 (11.1%) | 2 (22.2%) | 27 (300.0%) | 3 (33.3%) | 30 (333.3%) | 32 (355.6%) |
| sample_343 | 25 | 1075 | 13 (52.0%) | 5 (20.0%) | 2 (8.0%) | 5 (20.0%) | 9 (36.0%) | 7 (28.0%) | 16 (64.0%) | 21 (84.0%) |
| sample_365 | 57 | 1312 | 38 (66.7%) | 8 (14.0%) | 5 (8.8%) | 7 (12.3%) | 11 (19.3%) | 12 (21.1%) | 23 (40.4%) | 31 (54.4%) |
| sample_377 | 55 | 1420 | 20 (36.4%) | 9 (16.4%) | 10 (18.2%) | 17 (30.9%) | 7 (12.7%) | 27 (49.1%) | 34 (61.8%) | 43 (78.2%) |
| sample_378 | 24 | 939 | 11 (45.8%) | 3 (12.5%) | 0 (0.0%) | 10 (41.7%) | 1 (4.2%) | 10 (41.7%) | 11 (45.8%) | 14 (58.3%) |
| sample_380 | 50 | 1139 | 23 (46.0%) | 1 (2.0%) | 1 (2.0%) | 25 (50.0%) | 2 (4.0%) | 26 (52.0%) | 28 (56.0%) | 29 (58.0%) |
| sample_388 | 17 | 620 | 3 (17.6%) | 5 (29.4%) | 3 (17.6%) | 7 (41.2%) | 11 (64.7%) | 10 (58.8%) | 21 (123.5%) | 26 (152.9%) |
| sample_391 | 20 | 918 | 6 (30.0%) | 4 (20.0%) | 5 (25.0%) | 6 (30.0%) | 8 (40.0%) | 11 (55.0%) | 19 (95.0%) | 23 (115.0%) |
| sample_392 | 34 | 1004 | 11 (32.4%) | 5 (14.7%) | 3 (8.8%) | 17 (50.0%) | 3 (8.8%) | 20 (58.8%) | 23 (67.6%) | 28 (82.4%) |
| sample_393 | 50 | 1552 | 32 (64.0%) | 3 (6.0%) | 1 (2.0%) | 14 (28.0%) | 8 (16.0%) | 15 (30.0%) | 23 (46.0%) | 26 (52.0%) |
| sample_394 | 19 | 676 | 7 (36.8%) | 2 (10.5%) | 4 (21.1%) | 7 (36.8%) | 3 (15.8%) | 11 (57.9%) | 14 (73.7%) | 16 (84.2%) |
| sample_398 | 24 | 694 | 19 (79.2%) | 4 (16.7%) | 0 (0.0%) | 2 (8.3%) | 2 (8.3%) | 2 (8.3%) | 4 (16.7%) | 8 (33.3%) |
| sample_402 | 15 | 491 | 9 (60.0%) | 3 (20.0%) | 3 (20.0%) | 0 (0.0%) | 3 (20.0%) | 3 (20.0%) | 6 (40.0%) | 9 (60.0%) |
| sample_403 | 17 | 562 | 13 (76.5%) | 2 (11.8%) | 1 (5.9%) | 1 (5.9%) | 3 (17.6%) | 2 (11.8%) | 5 (29.4%) | 7 (41.2%) |
| sample_439 | 16 | 411 | 6 (37.5%) | 1 (6.3%) | 1 (6.3%) | 8 (50.0%) | 2 (12.5%) | 9 (56.3%) | 11 (68.8%) | 12 (75.0%) |
| sample_452 | 15 | 398 | 7 (46.7%) | 2 (13.3%) | 3 (20.0%) | 4 (26.7%) | 2 (13.3%) | 7 (46.7%) | 9 (60.0%) | 11 (73.3%) |
| sample_476 | 17 | 666 | 6 (35.3%) | 7 (41.2%) | 1 (5.9%) | 3 (17.6%) | 4 (23.5%) | 4 (23.5%) | 8 (47.1%) | 15 (88.2%) |
| sample_485 | 24 | 742 | 9 (37.5%) | 2 (8.3%) | 4 (16.7%) | 10 (41.7%) | 8 (33.3%) | 14 (58.3%) | 22 (91.7%) | 24 (100.0%) |
| sample_570 | 31 | 1038 | 15 (48.4%) | 12 (38.7%) | 3 (9.7%) | 4 (12.9%) | 15 (48.4%) | 7 (22.6%) | 22 (71.0%) | 34 (109.7%) |
| sample_579 | 30 | 932 | 18 (60.0%) | 4 (13.3%) | 2 (6.7%) | 6 (20.0%) | 5 (16.7%) | 8 (26.7%) | 13 (43.3%) | 17 (56.7%) |
| sample_583 | 29 | 927 | 19 (65.5%) | 3 (10.3%) | 1 (3.4%) | 6 (20.7%) | 7 (24.1%) | 7 (24.1%) | 14 (48.3%) | 17 (58.6%) |
| sample_664 | 19 | 885 | 13 (68.4%) | 1 (5.3%) | 4 (21.1%) | 1 (5.3%) | 2 (10.5%) | 5 (26.3%) | 7 (36.8%) | 8 (42.1%) |
| sample_666 | 40 | 873 | 23 (57.5%) | 7 (17.5%) | 2 (5.0%) | 11 (27.5%) | 3 (7.5%) | 13 (32.5%) | 16 (40.0%) | 23 (57.5%) |
| sample_687 | 20 | 426 | 7 (35.0%) | 8 (40.0%) | 1 (5.0%) | 7 (35.0%) | 0 (0.0%) | 8 (40.0%) | 8 (40.0%) | 16 (80.0%) |
| sample_70 | 7 | 519 | 5 (71.4%) | 1 (14.3%) | 0 (0.0%) | 1 (14.3%) | 1 (14.3%) | 1 (14.3%) | 2 (28.6%) | 3 (42.9%) |
| sample_71 | 12 | 562 | 3 (25.0%) | 6 (50.0%) | 0 (0.0%) | 5 (41.7%) | 3 (25.0%) | 5 (41.7%) | 8 (66.7%) | 14 (116.7%) |
| sample_782 | 10 | 355 | 3 (30.0%) | 1 (10.0%) | 0 (0.0%) | 6 (60.0%) | 0 (0.0%) | 6 (60.0%) | 6 (60.0%) | 7 (70.0%) |
| sample_930 | 32 | 853 | 27 (84.4%) | 0 (0.0%) | 2 (6.3%) | 3 (9.4%) | 6 (18.8%) | 5 (15.6%) | 11 (34.4%) | 11 (34.4%) |

| | | | | | | | | | |
|---|---|---|---|---|---|---|---|---|---|
| sample_942 | 10 | 285 | 8 (80.0%) | 0 (0.0%) | 1 (10.0%) | 1 (10.0%) | 1 (10.0%) | 2 (20.0%) | 3 (30.0%) | 3 (30.0%) |
| sample_945 | 24 | 766 | 15 (62.5%) | 3 (12.5%) | 2 (8.3%) | 4 (16.7%) | 7 (29.2%) | 6 (25.0%) | 13 (54.2%) | 16 (66.7%) |
| sample_96 | 23 | 759 | 13 (56.5%) | 8 (34.8%) | 0 (0.0%) | 4 (17.4%) | 2 (8.7%) | 4 (17.4%) | 6 (26.1%) | 14 (60.9%) |
| Correlation on counts | | | 0.66205 | 0.35365 | 0.29951 | 0.28844 | 0.51715 | 0.35374 | 0.61549 | 0.64534 |
| p-value on counts | | | 0.00000 | 0.00518 | 0.01903 | 0.02418 | 0.00002 | 0.00516 | 0.00000 | 0.00000 |
| Correlation on rates | | | 0.23061 | -0.10696 | -0.08415 | -0.15912 | 0.11767 | -0.19955 | 0.03385 | 0.00250 |
| p-value on rates | | | 0.07377 | 0.41190 | 0.51910 | 0.22060 | 0.36640 | 0.12310 | 0.79560 | 0.98480 |

Table S4: The occurrences of MMUD for entities inside and outside of dedicated sections with headers.

| | | GatorTron | | | GatorTron+CLAMP | | | GatorTronS | | | GatorTronS+CLAMP | | |
|---|---|---|---|---|---|---|---|---|---|---|---|---|---|
| Entity | Match | In | Out | p-value | In | Out | p-value | In | Out | p-value | In | Out | p-value |
| CC | EM+RM | 40 | 40 | 0.009 | 44 | 40 | 0.000 | 39 | 39 | 0.008 | 47 | 42 | 0.000 |
| | MMUD | 14 | 40 | | 10 | 40 | | 14 | 41 | | 7 | 39 | |
| HPI | EM+RM | 268 | 79 | 0.945 | 243 | 80 | 0.475 | 237 | 72 | 0.872 | 214 | 62 | 1.000 |
| | MMUD | 227 | 69 | | 254 | 72 | | 259 | 75 | | 279 | 82 | |
| Fam. H. | EM+RM | 40 | 4 | 0.208 | 40 | 3 | 0.172 | 39 | 2 | 0.006 | 39 | 2 | 0.034 |
| | MMUD | 1 | 1 | | 1 | 1 | | 2 | 3 | | 2 | 2 | |
| Past H. | EM+RM | 280 | 108 | 0.000 | 283 | 103 | 0.000 | 282 | 99 | 0.000 | 285 | 106 | 0.000 |
| | MMUD | 75 | 5 | | 73 | 61 | | 73 | 65 | | 71 | 58 | |
| Social H. | EM+RM | 96 | 20 | 0.000 | 97 | 19 | 0.000 | 97 | 22 | 0.000 | 90 | 15 | 0.000 |
| | MMUD | 9 | 14 | | 11 | 15 | | 9 | 12 | | 14 | 19 | |